\pdfoutput=1

\documentclass[11pt]{article}

\usepackage[preprint]{acl}

\usepackage{times}
\usepackage{latexsym}

\usepackage[T1]{fontenc}

\usepackage[utf8]{inputenc}

\usepackage{microtype}

\usepackage{inconsolata}

\usepackage{graphicx}

\usepackage{url}            
\usepackage{booktabs}       
\usepackage{amsfonts}       
\usepackage{nicefrac}       
\usepackage{xcolor}         
\usepackage{comment}
\usepackage{multirow}
\usepackage{xspace}
\usepackage{subcaption}
\usepackage{pifont}
\usepackage{ragged2e}
\usepackage{enumitem}
\usepackage{longtable}
\usepackage{color, colortbl}

\def\eg{{\em e.g.,}\xspace}
\def\vs{{\em v.s.}\xspace}
\def\ie{{\em i.e.,}\xspace}

\usepackage{textcomp}

\newcommand*{\yoruba}{Yor\`ub\'a\xspace}

\newcommand*{\flan}{FlanT5-XXL\xspace}
\newcommand*{\mto}{mT0-XXL-MT\xspace}
\newcommand*{\aya}{Aya-101\xspace}
\newcommand*{\bloom}{BLOOMZ 7B \xspace}
\newcommand*{\gemma}{Gemma 7B\xspace}
\newcommand*{\gemmaTwoL}{Gemma 2 27B\xspace}
\newcommand*{\llama}{LLaMa 2 7B\xspace}
\newcommand*{\llamas}{LLaMa 3.1 8B\xspace}
\newcommand*{\llamal}{LLaMa 3.1 70B\xspace}
\newcommand*{\commandr}{Command R\xspace}
\newcommand*{\commandrplus}{Command R+\xspace}
\newcommand*{\claude}{Claude Opus\xspace}
\newcommand*{\gptt}{GPT-3.5 Turbo\xspace}
\newcommand*{\gptturbo}{GPT-4-Turbo\xspace}
\newcommand*{\gpto}{GPT-4o\xspace}
\newcommand*{\gemini}{Gemini-1.5-Pro\xspace}
\newcommand*{\iroko}{\textsc{IrokoBench}\xspace}

\newcommand*{\anli}{AfriXNLI\xspace}
\newcommand*{\amath}{AfriMGSM\xspace}
\newcommand*{\ammlu}{AfriMMLU\xspace}

\definecolor{Gray}{gray}{0.85}
\definecolor{LightCyan}{rgb}{0.88,1,1}
\newcolumntype{a}{>{\columncolor{LightCyan}}r}

\title{IrokoBench: A New Benchmark for African Languages in the Age of Large Language Models}

\author{First Author \\
  Affiliation / Address line 1 \\
  Affiliation / Address line 2 \\
  Affiliation / Address line 3 \\
  \texttt{email@domain} \\\And
  Second Author \\
  Affiliation / Address line 1 \\
  Affiliation / Address line 2 \\
  Affiliation / Address line 3 \\
  \texttt{email@domain} \\}

\author{%
 David Ifeoluwa Adelani$^{1,2*}$, Jessica Ojo$^{1,3*}$, Israel Abebe Azime$^{4*}$, Jian Yun Zhuang$^{5}$, \\
 \textbf{Jesujoba O. Alabi$^{4}$, Xuanli He$^{6}$, Millicent Ochieng$^{7}$, Sara Hooker$^{8}$,  Andiswa Bukula$^{9}$,} \\ 
 \textbf{En-Shiun Annie Lee$^{10}$, Chiamaka Chukwuneke$^{11}$, Happy Buzaaba$^{12}$, Blessing Sibanda$^{*}$,} \\
 \textbf{Godson Kalipe$^{*}$, Jonathan Mukiibi$^{13*}$, Salomon Kabongo$^{14*}$, Foutse Yuehgoh$^{15*}$,} \\
 \textbf{Mmasibidi Setaka$^{9}$, Lolwethu Ndolela$^{*}$, Nkiruka Odu$^{*}$, Rooweither Mabuya$^{9}$, } \\ \textbf{Shamsuddeen Hassan Muhammad$^{16}$, Salomey Osei$^{17*}$, Sokhar Samb$^{18*}$,} \\
 \textbf{Tadesse Kebede Guge$^{19*}$, Tombekai Vangoni Sherman$^{20}$,  Pontus Stenetorp$^{6}$} \\\\
\footnotesize
$^*$Masakhane NLP, 
$^1$Mila, McGill University, $^2$Canada CIFAR AI Chair, 
$^3$Lelapa AI,
$^4$Saarland University, \\
\footnotesize
$^5$University of Toronto, 
$^6$University College London, 
$^7$Microsoft Research Africa,
$^8$Cohere For AI, 
$^9$SADiLaR, \\
\footnotesize
$^{10}$Ontario Tech University, 
$^{11}$Lancaster University, 
$^{12}$Princeton university, 
$^{13}$Makerere University, \\
\footnotesize
$^{14}$Leibniz Universität Hannover,
$^{15}$Le CNAM,
$^{16}$Imperial College London,
$^{17}$Universidad de Deusto,
$^{18}$DAUST, \\
\footnotesize
$^{19}$Haramaya University.\\
}

\begin{document}
\maketitle
\begin{abstract}
Despite the widespread adoption of Large language models (LLMs), their remarkable capabilities remain limited to a few high-resource languages. Additionally, many low-resource languages (\eg African languages) are often evaluated only on basic text classification tasks due to the lack of appropriate or 
comprehensive benchmarks outside of high-resource languages. In this paper, we introduce IrokoBench---a human-translated benchmark dataset for 17 typologically-diverse low-resource African languages covering three tasks: natural language inference~(\anli), mathematical reasoning~(\amath), and multi-choice knowledge-based question answering~(\ammlu). We use IrokoBench to evaluate zero-shot, few-shot, and translate-test settings~(where test sets are translated into English) across 10 open and six proprietary LLMs. Our evaluation reveals a significant performance gap between high-resource languages~(such as English and French) and low-resource African languages. We observe a significant performance gap between open and proprietary models, with the highest performing open model, \gemmaTwoL only at 63\% of the best-performing proprietary model GPT-4o performance. In addition, machine translating the test set to English before evaluation helped to close the gap for larger models that are English-centric, such as \gemmaTwoL and LLaMa 3.1 70B. These findings suggest that more efforts are needed to develop and adapt LLMs for African languages.
\end{abstract}

\section{Introduction}
In recent years, the capabilities of large language models~(LLMs) have greatly improved, from coherent chat experiences to solving complex and knowledge-intensive tasks like mathematical reasoning, coding, and question answering (QA)~\citep{openai2024gpt4,jiang2024mixtral,geminiteam2024gemini}. 
These models have also demonstrated the ability to quickly learn new and challenging tasks with few in-context learning examples and through chain-of-thought reasoning~\citep{brown2020language,shi2022language,wei2023chainofthought}. However, most state-of-the-art (SoTA) LLMs are primarily trained on high-resource languages (HRLs), resulting in sub-optimal performance for languages unseen during pre-training~\cite{touvron2023llama,ojo2023good}. Furthermore, 
this language coverage bias is reflected in the evaluation stage, predominantly conducted in English and a few other HRLs. 

There has been considerable effort to create benchmarks for African languages, but they typically cover simpler tasks, or are specific to narrow tasks such as machine translation, and, more recently, reading comprehension~\citep{bandarkar2023belebele,aremu2023yorc}.  Some diverse reasoning benchmarks have included Swahili---the most spoken native African language---for tasks like commonsense reasoning~\citep{ponti-etal-2020-xcopa} and natural language inference~\citep{conneau-etal-2018-xnli}. 
Consequently, current multilingual evaluations of LLMs do not accurately reflect their capabilities in reasoning and knowledge-intensive tasks across the majority of African languages.

Furthermore, the few comprehensive evaluations that exist across languages often rely on machine translation of English benchmarks \citep{singh2024aya}. While automatic translation from English benchmarks is a popular approach given the cost and time investment required for human translation, it often suffers from noise and biases 
~\citep{vanmassenhove-etal-2021-machine, lee-etal-2022-pre, khiu-etal-2024-predicting, hartung2023measuring, savoldi2021gender} or fail to reflect cultural context
~\citep{wang-etal-2022-measuring, ji_ji_bouillon_seligman_2023, pudjiati2022post}. 
Automatic curation may also amplify any of the ubiquitous issues with the quality of broad pretraining sets~\citep{luccioni-viviano-2021-whats,kreutzer-etal-2022-quality, ferrara2023should}. 

\textbf{In this paper, we seek to address both the \emph{diversity and breadth} of evaluation coverage.}  We introduce \iroko, a human curated benchmark dataset for 17 typologically diverse African languages which encompasses three complex tasks: natural language inference~(NLI), mathematical reasoning, and multi-choice knowledge-based QA. The datasets were created by human translating a subset of English cross-lingual NLI~(XNLI)~\citep{conneau-etal-2018-xnli}, English Multilingual Grade School Math~(MGSM)~\citep{shi2023language}, and Massive Multitask Language Understanding~(MMLU)~\citep{hendryckstest2021}, evaluation datasets into each of the 16 languages using professional translators.

We conduct a large-scale evaluation of \iroko to assess zero-shot, few-shot, and translate-test settings~(where test sets are translated into English) performance across 10 open and six proprietary LLMs. Our main contributions can be enumerated as follows:
\begin{enumerate}[leftmargin=15pt]
    \item \textbf{We introduce and release \iroko}, a human-translated benchmark that includes 16 languages from various geographical regions in Africa, all with varying degrees of ``low-resourcedness''~\citep{joshi-etal-2020-state}.  
    \vspace{-2mm}
    \item \textbf{Sharp cliff in performance across all models on low-resource languages} Our evaluation shows a large gap~(\scalebox{0.55}[1]{\(\sim\)}$45$\% on average) between the performance of high-resource languages~(\eg English) and African languages on all LLMs evaluated.
    Notably, Swahili performs better than other African languages, likely due  to its large corpus on the web. 
    \vspace{-2mm}
    \item \textbf{Models generally perform poorly in in-language evaluation} 
    This can be attributed to the inability of current SoTA LLMs to respond in the native languages of the users. Machine translating the test set to English before evaluation helped to close the gap for English-centric models; however, requiring users to always translate their prompts 
    to English may not be a desirable behavior.
    \vspace{-2mm}

    \item \textbf{\iroko highlights the performance divide between open and proprietary models on low-resource languages.} We find that proprietary closed models generally outperform open models for African languages. However, even these proprietary models exhibit substantial performance drops, due to the limited monolingual web data for African languages. The lowest performance is observed in languages such as \textit{Ewe}, \textit{Lingala}, \textit{Luganda}, \textit{Twi} and \textit{Wolof}, which each have less than 50 million characters of available data~\citep{kudugunta2023madlad400}. Among the tasks evaluated, \amath proves most challenging for LLMs, followed by \ammlu and \anli. 
\end{enumerate}

We will release \iroko on GitHub under the CC BY-SA 4.0 licence upon acceptance. 

\begin{figure*}[t]
    \centering
    \includegraphics[width=15cm]{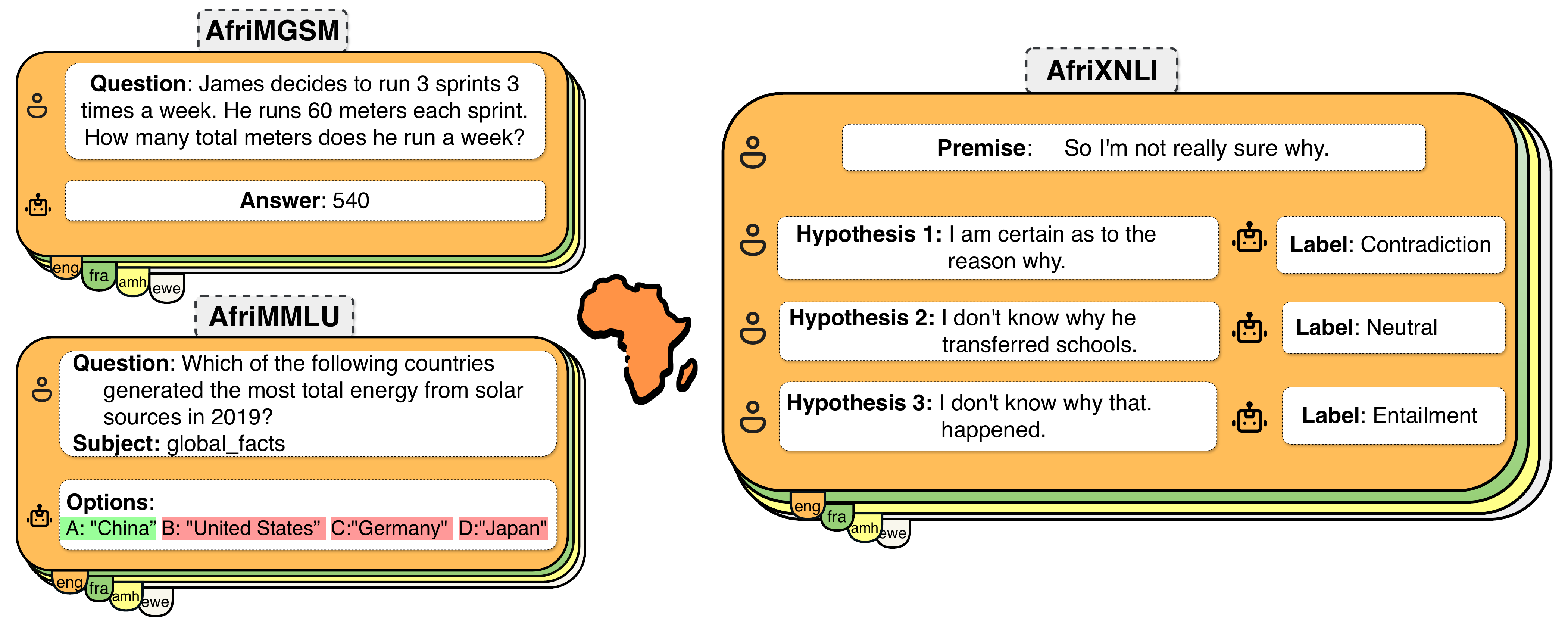}
    \caption{\textbf{Task Description for \iroko datasets.} 
    Both \amath and \ammlu focus on QA, while AfriXNLI focuses on natural language inference between two pairs of sentences. For clarity, this figure provides examples in  English.
    }
    \label{fig:tasks_image}
\end{figure*}

\section{Related Work}
\paragraph{Multilingual Evaluation of LLMs:}The evaluation of multilingual capabilities of LLMs has garnered significant attention. This has led to an increase in research that explores their performance across diverse linguistic landscapes \cite{ahuja-etal-2023-mega, Ahuja2023MEGAVERSEBL, lai-etal-2023-chatgpt, Hendy2023HowGA, bang-etal-2023-multitask,ustun2024aya,singh2024aya}. Despite this growing interest, there remains a notable lack of representation of African low-resource languages in these studies. \citet{ojo2023good,azime2024enhancing} address a broader spectrum of African languages, aligning more closely with our research. However, their study focuses on conventional NLP tasks, such as text classification, 
question answering, and text generation tasks. To address the lack of difficult benchmarks, a few works automatically translated MMLU benchmarks~\citep{lai-etal-2023-okapi}, but they do propagate errors of machine translation (MT) engines. Moreover, this is not applicable to low-resource languages with low-quality MT systems~\cite{adelani-etal-2022-thousand, nllbteam2022language}. Our research advances this by evaluating LLMs on more complex tasks using newly developed, human-annotated benchmarks specifically for African languages. 


\paragraph{African Benchmark Datasets:} Due to the limited representation of African languages in the field of NLP, there has been a growing effort to create benchmark datasets for African languages to enable research on these languages. 
Initiatives such as Masakhane
have been instrumental in the creation of standard benchmark for tasks such as machine translation \cite{adelani-etal-2022-thousand}, named entity recognition \cite{adelani-etal-2021-masakhaner,adelani-etal-2022-masakhaner}, part of speech tagging \cite{dione-etal-2023-masakhapos},  news topic classification \cite{adelani-etal-2023-masakhanews}, and sentiment analysis \cite{muhammad-etal-2023-afrisenti}. There are also several multilingual benchmark datasets that cover a few African languages, such as SIB-200 \cite{adelani-etal-2024-sib}, Flores \cite{goyal-etal-2022-flores,nllbteam2022language}, Aya dataset and Collection \citep{singh2024aya} and Taxi1500 \cite{ma2023taxi1500}. However, despite all these efforts, African languages still lack quality and more difficult datasets, our work fills this gap. 






\section{\iroko}
\subsection{Languages covered by \iroko}
\iroko\footnote{Our benchmark name comes from \`{I}r\'{o}k\`{o}--- a large hardwood in West Africa that is very durable for making bench.} cover 19 languages, 17 native African languages, and two European languages (English and French) which are widely spoken and official in many African countries. English and French are also the source languages we translated from. The 17 diverse and widely spoken African languages are from four regions of Africa: seven from West Africa (Ewe, Hausa, Igbo, Twi, Vai, Wolof, Yoruba), five from East Africa (Amharic, Kinyarwanda, Luganda, Swahili, and Oromo), four from Southern Africa (chiShona, isiXhosa, isiZulu, and Sesotho), and Central Africa (Lingala). These languages are from three language families in Africa: one from Mande, three from Afro-Asiatic and 13 from the Niger-Congo family---where we cover eight Bantu languages.  \autoref{tab:languages} provides an overview of the languages covered, including their family, regions, 
and the number of native speakers.

\subsection{Tasks covered by \iroko} \label{tasks} The selection of these tasks 
is primarily driven by their coverage across various domains and downstream tasks for diverse use cases. Additionally, they enable the evaluation of logical, abstract, and reasoning capabilities in LLMs, which is the hallmark of human intelligence~\cite{bowman-etal-2015-large,hendrycks2021measuring}.
\autoref{fig:tasks_image} provides examples of the different tasks covered in our datasets. We provide their descriptions below: 

\paragraph{\anli}The task of NLI involves the classification of a pair of sentences---a premise and a hypothesis as \textit{entailment}, \textit{neutral}, or \textit{contradiction} semantic relation. For example, the sentence ``so I'm not really sure why'' \textit{contradicts} ``I am certain as to the reason why'' but has a \textit{neutral} relation to ``I don't know why he transferred schools''. Here, we human translate the English portion of XNLI (a multilingual dataset comprising 15 languages, including Swahili) into the 15 African languages (excluding Swahili). 
While the original XNLI dataset has over 2,500/5,000 as DEV/TEST split, Each language in \anli has only 450 DEV instances and 600 test instances. We selected an equal number of instances from the 10 domains of XNLI. The task is evaluated using the \textit{accuracy metric} since the dataset has balanced classes. 

\begin{table*}[t] 
\centering
\resizebox{\textwidth}{!}{%
\begin{tabular}{llrp{90mm}p{28mm}}
\toprule
  & \textbf{No. of} & \textbf{No. instances}  &   & \textbf{Average Length} \\ 
  \textbf{Dataset} & \textbf{languages} & \textbf{Train/Dev/Test}  & \textbf{Subjects / Domains} & \textbf{Train/Dev/Test} \\ 

\midrule
\amath  & 17 (excl. Swahili, inc. Vai) & 8 \ / \ - \ / \ 250 & grade school mathematics & 25 \ / \ - \ / \ 46\\ 
\addlinespace
\ammlu  & 17 (incl. French)  & 25 \ / \ 83 \ / \ 500 & elementary mathematics, high-school geography, International law, global facts, high school microeconomics & 18 \ / \ 17 \ / \ 17\\
\addlinespace
\anli  & 15 (excl. Swahili)  &  - \ / \ 450 \ / \ 600 & face-to-face, telephone, oxford university press (oup), fiction, travel, government, nineeleven, letters, slate, verbatim & - \ /  10 \ /  10 (hyp.)  \  -  \ / 18 \ /  17 (pre.)\\
\bottomrule
\end{tabular}
}
\vspace{-2mm}
\caption{\textbf{The \iroko datasets:} dataset name, number of African languages covered, data split, and the subjects or domains covered. We included English, French, and Swahili in all benchmarks.}
\vspace{-2mm}
\label{tab:dataset}
\end{table*}

\paragraph{\ammlu} This is a multi-choice knowledge QA curated from freely available online sources by undergraduate and graduate students in the USA. The subjects cover simple general knowledge questions like ``global fact'' to highly-technical questions like ``professional law'' and ``professional medicine''. MMLU are often grouped into \textit{STEM}, \textit{Humanities}, \textit{Social Sciences}, and \textit{Others} category.  
We focus on five subjects that we believe 
are culturally unbiased (or international) and that are simpler to translate since many of the subjects covered are only taught in African countries using English or French, making it extremely difficult to translate highly technical subjects, especially the STEM subjects. \autoref{tab:dataset} shows the five subjects covered: two social science subjects (high-school geography and high-school microeconomics), one STEM subject (elementary mathematics), one humanities subject (international law), and one OTHER category (global facts). In total, we translated 608 question-answer pairs, with 500 instances in the test split, 100 questions per subject. The task is evaluated using the \textit{option prediction accuracy}.  

\paragraph{\amath} This is a QA task with questions obtained from grade school mathematical word problems created by human problem writers. \amath expands the original MGSM dataset~\citep{shi2023language}, which contains 250 QA pairs and 11 languages (including Swahili), to 15 more languages. The dataset consists of 8 training examples for few-shot and chain-of-thought prompting 
and 250 as a test set. We evaluated this task using the \textit{Exact Match metric}, which is popularly used for QA tasks. 


\subsection{Data collection process} 
\label{data collection}
\paragraph{Translation} We recruited language coordinators for each of the 16 African languages and French, and asked them to recruit professional translators to translate the sentences. The translation process took about two months, they started with XNLI, then MGSM and MMLU.  Each translator received an appropriate remuneration for their work.~\footnote{We recruited a logistic company in Kenya that managed all recruitment and payments---each country has different rates.  For example, we paid \$549.78 for the translation of 1020 XNLI samples in South Africa, \$355.86 in Nigeria. } Most of the translators translated from English except for \textit{Ewe}, \textit{Lingala} and \textit{Wolof} translators that translated from French since they are from the Francophone region of Africa. Additionally, we translated the MMLU dataset to French by professional translators and from French to these three languages. Many of the Francophone translators understand French and English but are more fluent in French, so they could cross-check from English if the French sentences were not clear enough.

\paragraph{Quality control} Regarding quality control, language coordinators reviewed and corrected any poorly translated sentences. Translators received payment only after this phase to ensure the quality of translations. For additional checks, we computed COMET~\citep{rei-etal-2020-comet} quality estimation (QE) scores between the human translation and the original sentences based on AfriCOMET QE metric~\citep{wang-etal-2024-afrimte}. In general, the distribution of the scores (range between 0 and 1) reflect that most translated sentences are between $0.7$ and $1.0$ for about 13 language pairs except for \textit{Lingala}, \textit{Twi}, and \textit{Wolof} where the average is around $0.5$. Further analysis shows that we cannot rely on these scores for those three languages since they are not covered in the pre-training of the original AfroXLMR encoder~\citep{alabi-etal-2022-adapting} used to build the AfriCOMET QE metric. Similar findings were reported in the original AfriCOMET QE paper that Twi had worse correlation with human judgement (\ie 0.279 for Pearson, and 0.060 for Spearman) \citep{wang-etal-2024-afrimte}.
We provide further analysis of the COMET scores in Appendix \ref{app:africomet_scores}.


\subsection{LLMs used for evaluation}
\label{models}

\paragraph{Open LLMs} We evaluate on two \emph{encoder-decoder open LLM}: mT0-XXL-MT~\citep{muennighoff-etal-2023-crosslingual}, and Aya-101~\citep{Ustun2024AyaMA} that have been instruction fine-tuned and multilingual T5 pre-trained on 101 languages (mT0 and Aya-101) models. Furthermore, these models are also all designed to be \emph{massively multilingual} and explicitly optimized to work outside of English. 
The languages covered during instruction tuning differ for different models, mT0 and Aya covered 46 and 101 languages respectively. 

Additionally we evaluate on eight \emph{decoder-only open LLM} models: BLOOMZ 7B~\citep{workshop2023bloom}, Gemma 2 (9B \& 27B)~\cite{Riviere2024Gemma2I},  LLaMa 3 8B~\citep{metaIntroducingMeta}, LLaMa 3.1 (8B \& 70B)~\citep{Dubey2024TheL3},
Command-R (August version)~\citep{cohereCommand}, and LLaMaX 3 8B~\citep{lu2024llamax}. 
These models' weights are openly available under various licenses, ranging from fully permissive to non-commercial, research-only licenses. We evaluate the instruction-tuned variant of these models. All models were pre-trained from scratch except LLaMaX that undergo continue pre-training on 100 languages including 13 languages covered in \iroko, except Ewe, Twi, Kinyarwanda and Vai. We used LLaMAX3-8B-Alpaca, instruction-tuned on English Alpaca~\cite{alpaca}.  




\paragraph{Closed LLM} We limit our evaluation to only OpenAI GPT  (3.5-0125, 4-Turbo-0125, 4o-mini-07-18, 4o-08-06)~\citep{chatgpt},  Gemini-1.5-Pro~\citep{Reid2024Gemini1U}, 
and Claude OPUS~\citep{anthropicClaude}
models. Recent work has shown that proprietary models tend to exhibit better multilingual capabilities~\cite{Ahuja2023MEGAVERSEBL}, although specifics regarding their pre-training and instruction fine-tuning processes are not disclosed.

\subsection{Evaluation Settings} 
\label{Evaluation Set-up}
\paragraph{Evaluation Set-up} We conduct two types of evaluations: \textit{in-language} and \textit{translate-test} evaluation, where test instances are automatically translated into English using a machine translation (MT) engine. For MT, we use NLLB-200 (3.3B)~\citep{nllbteam2022language}. In both in-language and translate-test setups, we perform cross-lingual transfer experiments from English and zero-shot evaluations by prompting LLMs. Few-shot evaluations are performed only for the three best models (two open and one closed) in the in-language setting. For AfriMGSM in both settings, we use \textbf{Chain-of-Thought (COT)} reasoning. 

We use the EleutherAI LM Evaluation Harness (\texttt{lm-eval}) tool~\citep{biderman2024lessons}---a popular evaluation tool that is helping to standardize LLM evaluation, especially for open models on HuggingFace Model Hub. For closed models, 
we employ a \textbf{verbalizer}~\cite{gao-etal-2021-making,schick2021s} for prediction and evaluation. All models are prompted with \textbf{five different templates}.  We provide more details in Appendix \ref{app:eval_tool}. \footnote{We make use of Cohere API for  Command-R inference.}

\paragraph{Cross-lingual transfer experiments} We first conduct a study on cross-lingual transfer in a supervised learning setting by fine-tuning the English training data ($400K$ instances) from ~\citet{conneau-etal-2018-xnli} and evaluating on the remaining languages. This experiment focuses solely on the NLI task due to the availability of 
training data for supervised learning. The evaluation employs several masked language models, including XLM-R~\cite{conneau-etal-2020-unsupervised}, Serengeti~\cite{adebara-etal-2023-serengeti}, AfroXLMR-\{base, large\}~\cite{alabi-etal-2022-adapting}, AfroXLMR-76L~\cite{adelani-etal-2024-sib}. We report the result of the best model in the paper and others in 
Appendix \ref{app:cross_tranf}.  

\paragraph{Zero- and few-shot evaluation} In a zero-shot setting, we use the prompts detailed in \autoref{app:eval_tool}. For few-shot evaluations, we conduct a 5-shot assessment for both \ammlu and \anli, and an 8-shot assessment for \amath.

\begin{table*}[!ht]
    \centering
    \resizebox{\textwidth}{!}{%
    \begin{tabular}{ll|rr|rr|rr|rrrr}
    \toprule
          ~ & ~ & \multicolumn{2}{c}{\textbf{\anli}} & \multicolumn{2}{c}{\textbf{\ammlu}} & \multicolumn{2}{c}{\textbf{\amath }} & \textbf{Ave.} & \textbf{Ave.} & \textbf{Ave.} & \textbf{Ave.} \\ 
         &  & \textbf{in-} &  \textbf{translate} & \textbf{in-} & \textbf{translate} & \textbf{in-} & \textbf{translate} & \textbf{in-} & \textbf{translate} & \textbf{English}  & \textbf{French} \\ 
        \textbf{Model} & \textbf{size} & \textbf{lang.} & \textbf{test} &  \textbf{lang.} & \textbf{test} &  \textbf{lang.} & \textbf{test} &  \textbf{lang.} & \textbf{test} & \textbf{lang.} & \textbf{lang.} \\ 
        \midrule
        AfroXLMR-76L & 559M & \textbf{65.7} & \textbf{63.6} & ~ & ~ & ~ & ~ & ~ & ~ & ~ & ~ \\ 
        \midrule
        mT0-XXL-MT & 13B & 51.0 & 49.9 & 27.9 & 28.4 & 2.9 & 2.5 & 27.3 & 26.9 & 34.5 & 33.1 \\ 
        Aya-101 & 13B & 51.5 & 50.2 & 29.7 & 31.1 & 4.6 & 7.9 & 28.6 & 29.7 & 39.3 & 35.3 \\
        \midrule
        BLOOMZ 7B & 7B & 39.4 & 47.6 & 24.1 & 27.9 & 1.7 & 1.9 & 21.7 & 25.8 & 31.4 & 28.9 \\ 
        LLaMa 3 8B & 8B & 35.4 & 38.2 & 28.1 & 31.8 & 3.9 & 33.4 & 22.5 & 34.4 & 52.8 & 45.8 \\ 
        LLaMa 3.1 8B & 8B & 36.6 & 43.6 & 31.1 & 41.1 & 7.2 & 30.1 & 24.9 & 38.2 & 57.5 & 53.4 \\ 
        LLaMaX 3 8B & 8B & 40.8 & 33.3 & 29.3 & 35.2 & 4.8 & 9.2 & 24.9 & 25.9 & 38.5 & 33.1 \\ 
        Gemma 2 9B & 9B & 40.3 & 43.3 & 35.4 & 44.7 & 19.8 & 39.4 & 31.9 & 42.5 & 64.6 & 58.2 \\ 
        \rowcolor{Gray}
        Gemma 2 27B & 27B & 42.8 & 49.0 & 39.9 & 48.8 & 28.5 & \textbf{46.1} & 37.1 & 48.0 & 76.3 & 69.9 \\ 
        \rowcolor{Gray}
        LLaMa 3.1 70B & 70B & 38.0 & 42.8 & 39.4 & 51.3 & 24.6 & 45.6 & 34.0 & 46.5 & 73.5 & 65.5 \\ 
        Command-R & 35B & 43.4 & \underline{57.0} & 27.8 & 40.8 & 5.7 & 38.3 & 25.6 & 45.4 & 71.7 & 63.0 \\
        \midrule
        Claude Opus & UNK & 58.1 & 56.4 & 43.0 & 47.6 & 25.3 & 32.7 & 42.3 & 45.6 & 73.3 & 64.0 \\ 
        Gemini-1.5-Pro & UNK & 59.4 & 49.9 & \textbf{60.2} & \underline{53.1} & \textbf{55.4} & \textbf{44.3} & \underline{58.3} &  49.1 & 82.6 & 75.3 \\ 
        GPT-3.5-Turbo & UNK & 42.1 & 45.5 & 38.1 & 46.8 & 10.6 & 37.1 & 30.2 & 43.1 & 71.9 & 62.9 \\ 
        GPT-4o-mini & UNK & 54.2 & 56.7 & 45.5 & 50.2 & 35.4 & 42.3 & 45.0 & 49.7 & \underline{84.7} & \underline{76.7} \\ 
         \rowcolor{Gray}
        GPT-4-Turbo & UNK & 59.5 & \underline{57.0} & 54.2 & 52.1 &  45.2 & \underline{43.5} & 52.9 & \textbf{50.9} & 84.6 & 76.5 \\ 
         \rowcolor{Gray}
        GPT-4o & UNK & \underline{64.3} & 52.1 & \underline{60.0} & \textbf{54.1} & \underline{52.6} & 42.6 & \textbf{59.0} & \underline{49.6} & \textbf{86.9} & \textbf{78.1} \\ 
        \midrule
        ~ & ~ & ~ & ~ & ~ & ~ & ~ & ~ & 33.9 & 40.2 & 62.8 & 56.3 \\ 
        \bottomrule
    \end{tabular}
    }
    \vspace{-2mm}
    \caption{\textbf{Main results:} Average performance of various LLMs on all tasks (ave. excl. eng, fra, and vai). Except for \amath, which uses the Exact Match metric, others use the Accuracy. The best result is in \textbf{Bold} and second best \underline{underlined}. The top-2 open and closed models are in \textbf{gray}. We report only the \textbf{best prompt} (others in appendix). } 
    \label{tab:main_results}
\end{table*}

\begin{table*}[!ht]
    \centering
    \resizebox{\textwidth}{!}{%
    \begin{tabular}{l|rrrrrrrrrrrrrrrrrr|r}
    \toprule
\textbf{Model} & \textbf{eng}  & \textbf{fra} & \textbf{amh}  & \textbf{ewe}  & \textbf{hau}  & \textbf{ibo}  & \textbf{kin}  & \textbf{lin}  & \textbf{lug}  & \textbf{orm}  & \textbf{sna}  & \textbf{sot}  & \textbf{swa}  & \textbf{twi}  & \textbf{wol}  & \textbf{xho}  & \textbf{yor}  & \textbf{zul}  & \textbf{ave}  \\
\midrule
    \multicolumn{9}{l}{\textit{Prompt LLMs in African Language}}    \\
        \aya \texttt{(t2)} & 40.0 & 36.6 & 31.6 & 25.4 & 33.4 & 36.8 & 30.8 & 27.8 & 28.0 & 26.2 & 28.2 & 31.8 & 32.2 & 26.8 & 25.2 & 32.0 & 28.4 & 29.8 & 29.7 \\ 
        \gemmaTwoL \texttt{(t1)} & 75.6 & 66.4 & 40.6 & 32.4 & 43.2 & 44.2 & 40.2 & 38.2 & 32.6 & 33.6 & 44.6 & 41.8 & 56.0 & 35.6 & 30.4 & 42.0 & 41.0 & 42.2 & 39.9 \\ 
        \llamal \texttt{(t1)} & 76.4 & 69.4 & 41.6 & 32.2 & 47.6 & 47.2 & 38.6 & 40.0 & 34.4 & 35.6 & 41.6 & 39.0 & 55.8 & 28.4 & 31.6 & 34.2 & 41.4 & 40.4 & 39.4 \\ 
        \gpto \texttt{(t1)} & \textbf{87.4} & \textbf{83.2} & \textbf{59.8} & \textbf{33.6} & \textbf{67.2} & \textbf{67.2} & \textbf{64.2} & \textbf{61.0} & \textbf{52.8} & \textbf{61.0} & \textbf{67.6} & \textbf{67.4} & \textbf{77.4} & \textbf{43.2} & \textbf{37.8} & \textbf{70.2} & \textbf{61.2} & \textbf{68.2} & \textbf{60.0} \\ 
        \midrule
        \multicolumn{9}{l}{\textit{Translate-Test (Eval. in English)}}    \\
        \aya \texttt{(t2)} & ~ & 37.8 & 32.4 & 28.6 & 31.0 & 31.6 & 31.8 & 33.6 & 27.2 & 28.6 & 32.4 & 34.2 & 31.2 & 31.8 & 26.6 & 30.8 & 34.2 & 32.2 & 31.1 \\ 
        \gemmaTwoL \texttt{(t1)} & ~ & 62.4 & 57.6 & 40.4 & 50.0 & 50.0 & 50.6 & 47.0 & 42.8 & 46.6 & 49.8 & 55.6 & 59.8 & 39.8 & 31.2 & 55.2 & 52.6 & 51.6 & 48.8 \\ 
        \llamal \texttt{(t1)} & ~ & 67.4 & 55.6 & \textbf{44.8} & 50.6 & 55.8 & 55.6 & 53.8 & \textbf{46.8} & 49.4 & 53.6 & 59.6 & 63.0 & 41.2 & \textbf{32.8} & 55.4 & 49.6 & 52.8 & 51.3 \\ 
        \gpto \texttt{(t1)} & ~ & \textbf{76.4} & \textbf{62.8} & 43.8 & \textbf{54.0} & \textbf{57.4} & \textbf{58.2} & \textbf{54.4} &  46.4 & \textbf{54.8} & \textbf{55.6} & \textbf{63.2} & \textbf{67.4} & \textbf{44.0} & 32.0 & \textbf{62.2} & \textbf{52.4} & \textbf{57.4} & \textbf{54.1} \\ 
    \bottomrule
    \end{tabular}
    }
    \vspace{-2mm}
    \caption{\textbf{\ammlu results in in-language and translate-test scenarios}: Option prediction accuracy per language. Average computed on only African languages. The \textbf{best prompt template} for each model is in \textbf{bracket}. }
     \vspace{-2mm}
\label{tab:afrimmlu_table}
\end{table*}

\section{Results}
\subsection{Overall Results}

\paragraph{Large performance gaps between high-resource languages and African languages}  \autoref{tab:main_results} shows the result of zero-shot evaluation for various LLMs. On average, there is a significant performance gap between African languages and English (up to $28\%$) and French (up to $19\%$) on the best LLM. The best LLM for African languages is \gpto, with an average performance of 59.0 across the evaluated tasks. 
Finally, as shown in \autoref{tab:languages}, the languages with the lowest performance have the least monolingual data 
on the web.

\paragraph{Large performance gaps between closed and open weights models} Our results, as presented in \autoref{tab:main_results}, indicate that the closed models \claude, \gemini, \gptturbo, and \gpto consistently outperform the open models on \iroko. The top-2 closed models achieve average performance scores ranging from $52.9$ to $59.0$ across all tasks. The performance gap between the best closed model (\gpto) and the best open model (\gemmaTwoL) is $21.9$. Notably, the largest performance differences for \textit{in-language} setting are observed in the \ammlu and \amath tasks, where \gpto outperforms \gemmaTwoL by $20.1$ and $24.1$, respectively. For the \anli task, \mto and \aya perform better than bigger models like \gemmaTwoL and \llamal with 27B and 70B parameters respectively.



\paragraph{Majority of models perform worse for in-language prompting} Most users would prefer to prompt in their native language; however, we find that almost all models we benchmark perform better with prompts translated into English. Only a few exceptions, such as \gptturbo, and \gpto, perform better in the in-language evaluation. 
Specifically, \commandr and \llamas benefit the most from the translate-test approach, showing average improvements of $+19.8$ and $+13.3$, respectively. Notably, \gemmaTwoL achieves the best overall results for the \amath task with the translate-test, outperforming \gpto by $+3.5$. 
We attribute this boost to the fact that these models are heavily English-centric. 


\begin{table*}[!ht]
    \centering
     \resizebox{\textwidth}{!}{%
    \begin{tabular}{p{40mm}|aarrarrrrrraarrara|r}
    \toprule
\textbf{Model} & \textbf{eng}  & \textbf{fra} & \textbf{amh}  & \textbf{ewe}  & \textbf{hau}  & \textbf{ibo}  & \textbf{kin}  & \textbf{lin}  & \textbf{lug}  & \textbf{orm}  & \textbf{sna}  & \textbf{sot}  & \textbf{swa}  & \textbf{twi}  & \textbf{wol}  & \textbf{xho}  & \textbf{yor}  & \textbf{zul}  & \textbf{ave}  \\
\midrule
        Elementary Mathematics & 88.0 & 83.0 & 76.0 & 55.0 & 73.0 & 84.0 & 77.0 & 71.0 & 80.0 & 81.0 & 75.0 & 78.0 & 87.0 & 68.0 & 61.0 & 76.0 & 80.0 & 80.0 & 75.1 \\ 
        Global Facts & 67.0 & 60.0 & 44.0 & 32.0 & 56.0 & 49.0 & 60.0 & 54.0 & 41.0 & 52.0 & 48.0 & 49.0 & 59.0 & 33.0 & 36.0 & 56.0 & 42.0 & 57.0 & 48.0 \\ 
        High School Geography & 93.0 & 86.0 & 56.0 & 25.0 & 67.0 & 64.0 & 57.0 & 55.0 & 48.0 & 54.0 & 69.0 & 68.0 & 76.0 & 34.0 & 27.0 & 63.0 & 53.0 & 73.0 & 55.6 \\ 
        High School Microeconomics & 98.0 & 90.0 & 54.0 & 26.0 & 70.0 & 58.0 & 55.0 & 63.0 & 44.0 & 55.0 & 59.0 & 62.0 & 82.0 & 32.0 & 19.0 & 78.0 & 51.0 & 61.0 & 54.3 \\ 
        International Law & 90.0 & 91.0 & 64.0 & 36.0 & 71.0 & 75.0 & 75.0 & 76.0 & 55.0 & 65.0 & 78.0 & 72.0 & 86.0 & 45.0 & 38.0 & 84.0 & 75.0 & 76.0 & 66.9 \\ 
        \midrule
        Average & 87.2 & 82.0 & 58.8 & 34.8 & 67.4 & 66.0 & 64.8 & 63.8 & 53.6 & 61.4 & 65.8 & 65.8 & 78.0 & 42.4 & 36.2 & 71.4 & 60.2 & 69.4 & 60.0 \\ 
     \bottomrule
    \end{tabular}
    }
    \vspace{-2mm}
    \caption{\textbf{GPT-4o \ammlu results by subjects}: Option prediction accuracy per language. Languages with at least 60\% accuracy in four subjects are in \colorbox{LightCyan}{Cyan}.}
    \label{tab:mmlu_subjects}
\end{table*}

\begin{table*}[!ht]
    \centering
    \resizebox{\textwidth}{!}{%
    \begin{tabular}{l|rrrrrrrrrrrrrrrrrr|r}
    \toprule
\textbf{Model} & \textbf{eng}  & \textbf{fra} & \textbf{amh}  & \textbf{ewe}  & \textbf{hau}  & \textbf{ibo}  & \textbf{kin}  & \textbf{lin}  & \textbf{lug}  & \textbf{orm}  & \textbf{sna}  & \textbf{sot}  & \textbf{swa}  & \textbf{twi}  & \textbf{wol}  & \textbf{xho}  & \textbf{yor}  & \textbf{zul}  & \textbf{ave}  \\
\midrule
        \multicolumn{9}{l}{\textit{Prompt LLMs in African Language}}    \\
        \aya \texttt{(t1)} & 10.8 & 9.6 & 6.8 & 3.2 & 7.2 & 3.2 & 3.6 & 4.4 & 2.4 & 4.8 & 6.8 & 10.8 & 6.0 & 1.6 & 1.2 & 4.4 & 3.2 & 3.6 & 4.6 \\ 
        \gemmaTwoL \texttt{(t4)} & 85.6 & \textbf{80.0} & 33.6 & 7.6 & 49.6 & 24.0 & 32.4 & 18.0 & 23.6 & 12.8 & 35.2 & 38.4 & 73.6 & 12.4 & 5.6 & 32.0 & 22.4 & 34.4 & 28.5 \\ 
        \llamal \texttt{(t4)} & \textbf{86.8} & 76.4 & 17.6 & 8.0 & 48.8 & 37.2 & 26.4 & 11.2 & 24.4 & 10.8 & 21.2 & 32.8 & 68.0 & 14.4 & 3.2 & 18.8 & 23.6 & 27.2 & 24.6 \\ 
        \gpto \texttt{(t2)} & 84.0 & 68.8 & \textbf{57.6} & \textbf{8.8} & \textbf{64.8} & \textbf{57.6} & \textbf{60.4} & \textbf{51.2} & \textbf{51.6} & \textbf{61.2} & \textbf{58.4} & \textbf{60.8} & \textbf{78.8} & \textbf{31.2} & \textbf{28.0} & \textbf{52.4} & \textbf{62.0} & \textbf{57.2} & \textbf{52.6} \\ 
        \midrule
        \multicolumn{9}{l}{\textit{Translate-Test (Eval. in English)}}    \\
        \aya \texttt{(t1)} & ~ & 8.4 & 8.4 & 6.4 & 7.6 & 6.0 & 10.0 & 6.4 & 6.8 & 7.6 & 6.8 & 10.4 & 9.2 & 6.0 & 8.4 & 8.8 & 8.8 & 8.0 & 7.9 \\ 
        \gemmaTwoL \texttt{(t4)} & & 70.8 & 53.2 & 30.0 & \textbf{54.0} & \textbf{44.0} & \textbf{55.2} & 47.2 & 34.4 & \textbf{46.0} & \textbf{48.0} & 54.4 & 69.6 & \textbf{29.2} & \textbf{21.6} & 48.4 & \textbf{48.4} & \textbf{54.0} & \textbf{46.1} \\ 
        \llamal \texttt{(t4)} & & \textbf{73.6} & \textbf{54.8} & \textbf{30.4} & 48.0 & 43.2 & 52.8 & \textbf{48.0} & \textbf{35.6} & 44.0 & 46.8 & \textbf{55.2} & \textbf{72.0} & 26.4 & 20.4 & \textbf{49.6} & \textbf{48.4} & \textbf{54.0} & 45.6 \\ 
        \gpto  \texttt{(t2)} & & 70.0 & 48.4 & 23.6 & 46.8 & 39.2 & 51.6 & 44.8 & 35.2 & 42.0 & 42.4 & 54.8 & 68.0 & 26.4 & 16.0 & 45.6 & 47.2 & 49.2 & 42.6 \\ 
    \bottomrule
    \end{tabular}
    }
    \vspace{-2mm}
    \caption{\textbf{\amath results in in-language and translate-test scenarios}: Exact Match score per language. Average computed on only African languages.  The \textbf{best prompt template} for each model is in \textbf{bracket}. }
\label{tab:afrimgsm_detailed}
\end{table*}

\begin{table*}[!ht]
    \centering
    \resizebox{\textwidth}{!}{%
\begin{tabular}{l|rrrrrrrrrrrrrrrrrr|r}
\toprule
\textbf{Model} & \textbf{eng}  & \textbf{fra} & \textbf{amh}  & \textbf{ewe}  & \textbf{hau}  & \textbf{ibo}  & \textbf{kin}  & \textbf{lin}  & \textbf{lug}  & \textbf{orm}  & \textbf{sna}  & \textbf{sot}  & \textbf{swa}  & \textbf{twi}  & \textbf{wol}  & \textbf{xho}  & \textbf{yor}  & \textbf{zul}  & \textbf{ave}  \\
\midrule
\multicolumn{9}{l}{\textit{Prompt LLMs in African Language}}    \\
        AfroXLMR-76L  & 88.2 & \textbf{83.3} & \textbf{78.5} & \textbf{58.3} & \textbf{73.3} & \textbf{70.0} & \textbf{65.8} & \textbf{33.3} & \textbf{68.0} & \textbf{69.3} & \textbf{70.8} & \textbf{70.8} & \textbf{73.3} & \textbf{59.5} & \textbf{51.8} & \textbf{73.0} & \textbf{63.2} & \textbf{72.5} & \textbf{65.7} \\ 
        \aya (t4) & 67.0 & 59.7 & 64.2 & 43.2 & 57.0 & 55.5 & 54.3 & 33.5 & 51.7 & 51.5 & 55.7 & 52.2 & 56.5 & 47.0 & 36.7 & 55.2 & 54.5 & 55.3 & 51.5 \\ 
        \gemmaTwoL \texttt{(t2)} & 67.8 & 63.3 & 47.0 & 36.8 & 49.7 & 46.2 & 40.5 & 32.0 & 41.7 & 35.8 & 46.0 & 43.5 & 57.0 & 36.0 & 36.7 & 45.0 & 42.5 & 48.0 & 42.8 \\ 
        \llamal \texttt{(t2)} & 57.3 & 50.7 & 43.2 & 34.3 & 42.8 & 42.3 & 36.5 & 32.8 & 37.5 & 34.7 & 35.5 & 38.3 & 44.0 & 36.0 & 34.7 & 39.3 & 39.0 & 37.0 & 38.0 \\ 
        \gpto \texttt{(t3)} & \textbf{89.2} & 82.3 & 71.8 & 45.0 & 75.2 & 68.2 & 68.0 & 32.7 & 69.8 & 71.2 & 71.3 & 71.8 & 71.5 & 55.8 & 52.7 & 72.0 & 64.5 & 67.5 & 64.3 \\ 
        \midrule
        \multicolumn{9}{l}{\textit{Translate-Test (Eval. in English)}}    \\
        AfroXLMR-76L  & ~ & \textbf{83.0} & \textbf{73.7} & \textbf{54.3} & \textbf{67.2} & \textbf{66.0} & \textbf{63.0} & 32.8 & \textbf{65.7} & \textbf{65.8} & \textbf{71.2} & \textbf{70.2} & \textbf{73.0} & \textbf{56.8} & \textbf{47.5} & \textbf{74.2} & \textbf{63.7} & \textbf{72.0} & \textbf{63.6} \\ 
        \aya (t4) & ~ & 61.2 & 60.7 & 40.8 & 53.8 & 52.3 & 50.3 & \textbf{33.0} & 48.7 & 51.3 & 54.0 & 56.0 & 54.0 & 44.5 & 39.3 & 56.8 & 51.0 & 57.0 & 50.2 \\ 
        \gemmaTwoL \texttt{(t2)} & ~ & 60.8 & 55.8 & 45.5 & 48.5 & 49.7 & 47.7 & 31.5 & 51.3 & 52.3 & 51.7 & 50.2 & 55.7 & 47.2 & 40.2 & 54.8 & 50.2 & 51.5 & 49.0 \\ 
        \llamal \texttt{(t2)} & ~ & 58.8 & 42.8 & 39.3 & 45.2 & 39.8 & 36.2 & 32.7 & 44.3 & 46.0 & 51.8 & 50.8 & 52.5 & 38.8 & 34.5 & 44.8 & 41.8 & 42.8 & 42.8 \\ 
        \gpto \texttt{(t3)} & ~ & 73.8 & 63.8 & 42.8 & 54.8 & 53.3 & 52.8 & 32.3 & 54.3 & 55.5 & 58.2 & 58.0 & 58.5 & 45.7 & 38.8 & 58.2 & 53.5 & 52.8 & 52.1 \\ 
\bottomrule
    \end{tabular}
}
\vspace{-2mm}
\caption{\textbf{\anli results in in-language and translate-test scenarios}: Option prediction accuracy per language. Average computed on only African languages. The \textbf{best prompt template} for each model is in \textbf{bracket}.}
\label{tab:nli_results}

\end{table*}

\begin{figure*}[t]
\centering
    \begin{subfigure}{0.32\textwidth}
        \centering
\includegraphics[width=0.95\textwidth]{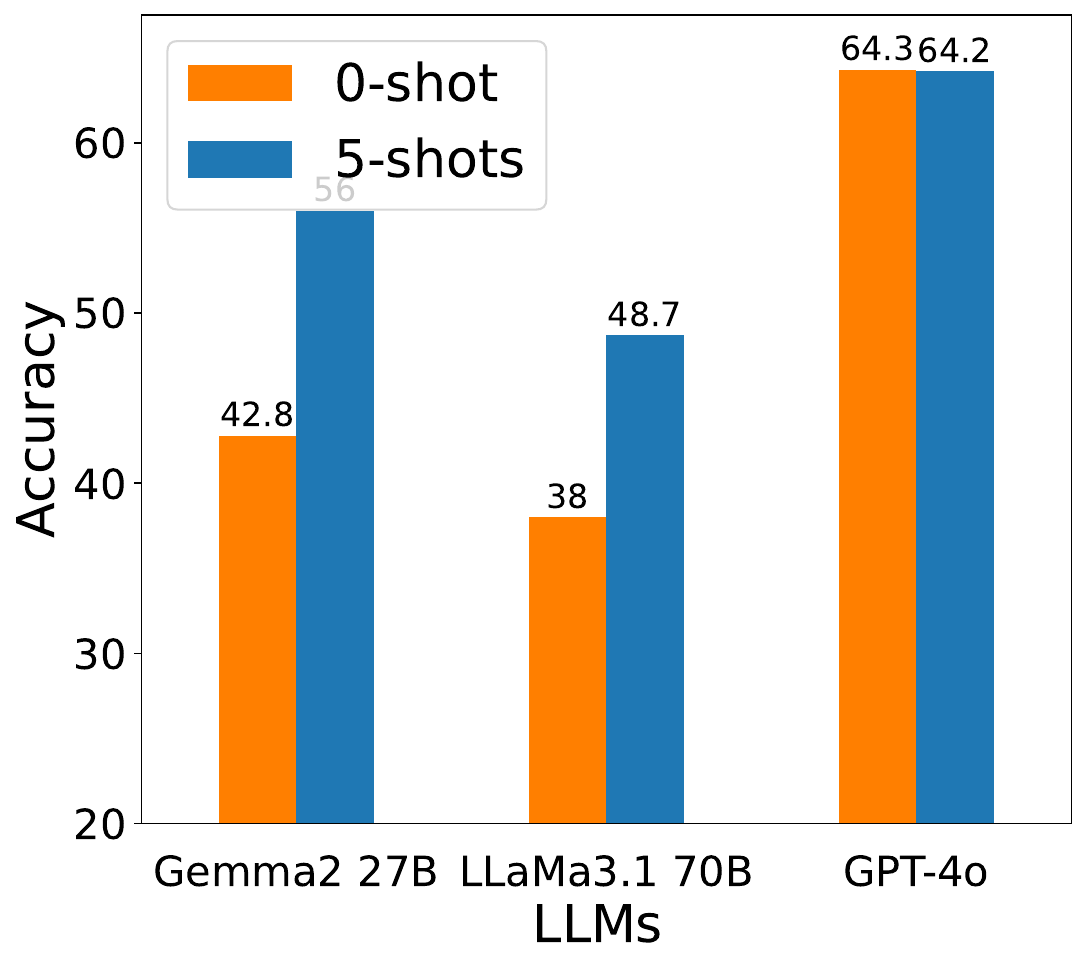}
        \caption{AfriXNLI} \label{fig:afrixnli}
    \end{subfigure}
    \begin{subfigure}{0.32\textwidth}
        \centering
    \includegraphics[width=0.95\textwidth]{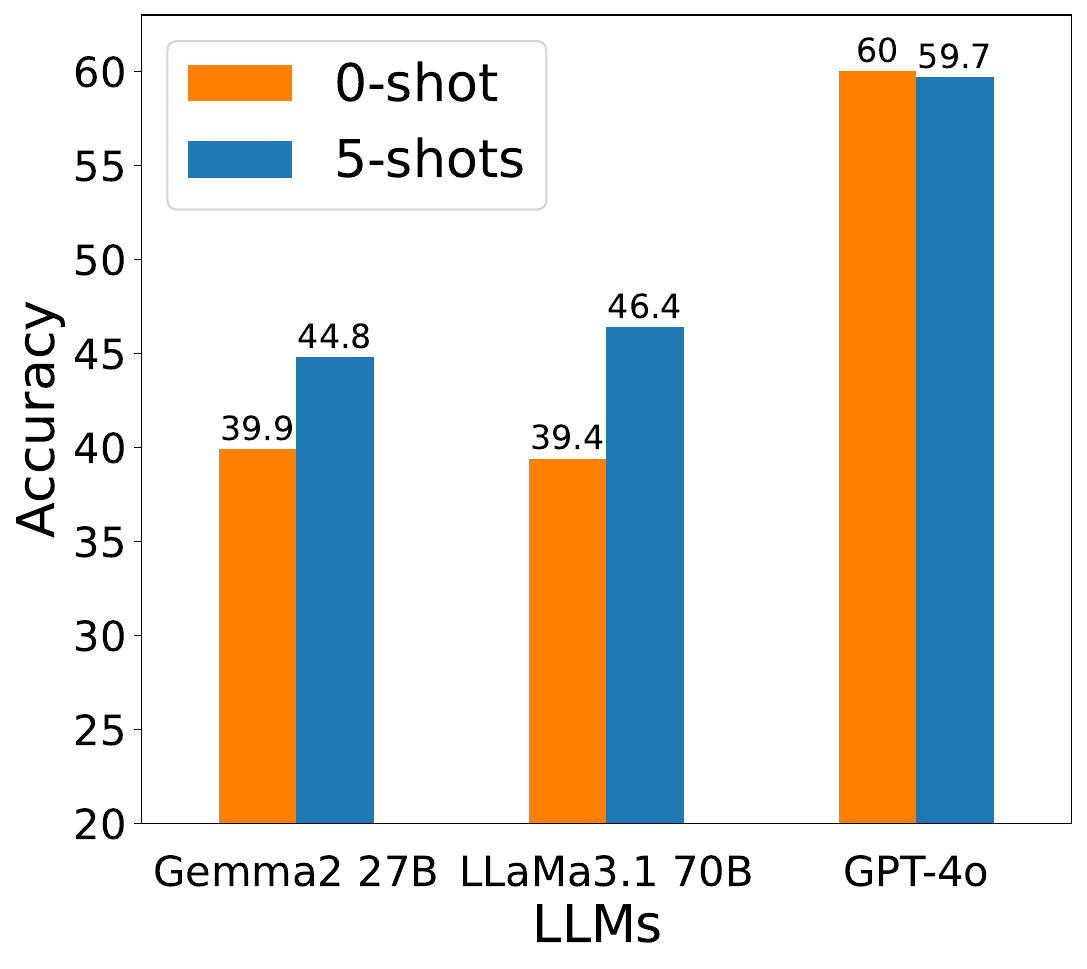}
        \caption{AfriMMLU} \label{fig:afrimmlu}
    \end{subfigure}
    \begin{subfigure}{0.32\textwidth}
        \centering
        \includegraphics[width=0.95\textwidth]{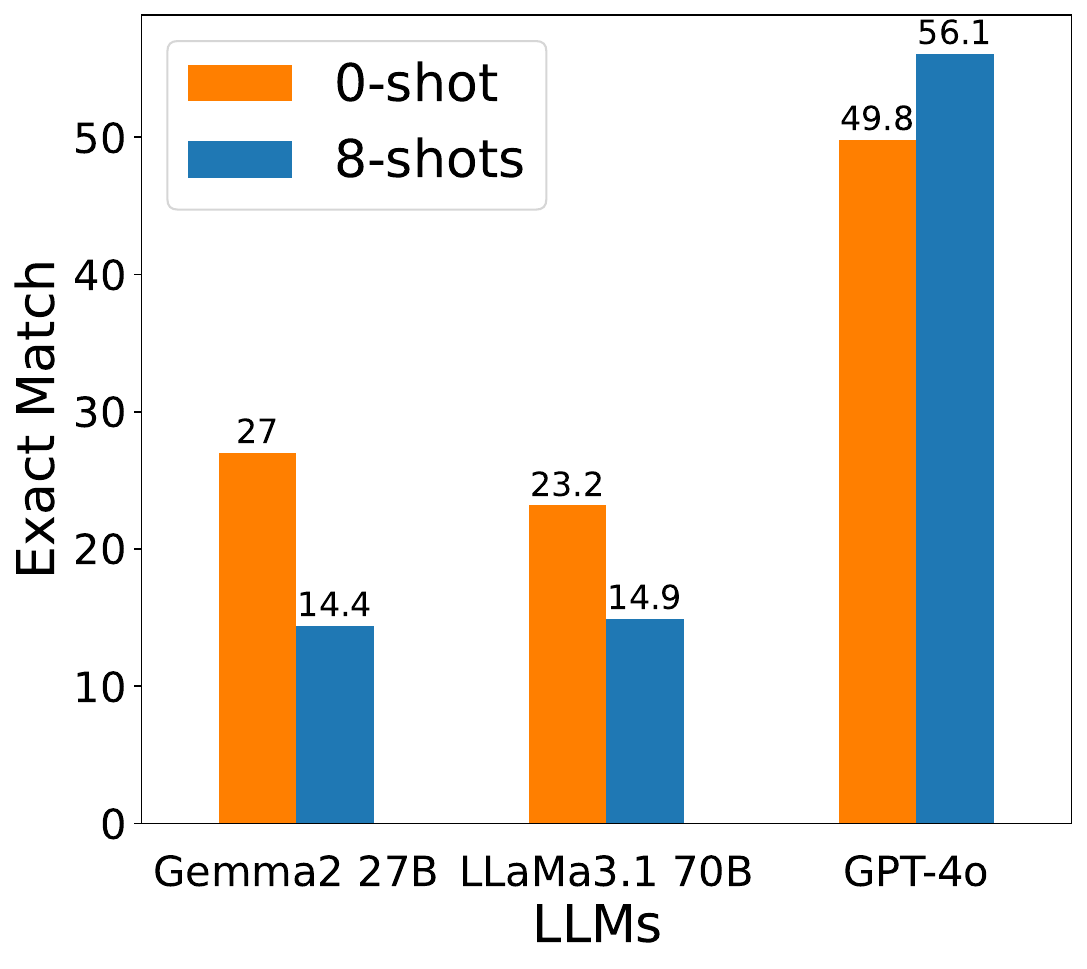}
        \caption{AfriMGSM} \label{fig:afrimgsm}
    \end{subfigure}
    \vspace{-3mm}
    \caption{\textbf{Few-shot evaluation} on \iroko, we performed 8-shots for \amath and 5-shots for the others.
    \vspace{-3mm}
    \label{fig:few_shot}}
\end{figure*}

\begin{table*}[t]
    \centering
     \resizebox{\textwidth}{!}{%
    \begin{tabular}{lrrrrrc|rrrrrc|rrrrrc}
    \toprule
        & \multicolumn{6}{c}{\anli}  &  \multicolumn{6}{c}{\ammlu}  & \multicolumn{6}{c}{\amath} \\ 
        \textbf{Model} & \textbf{t1} & \textbf{t2} & \textbf{t3} & \textbf{t4} & \textbf{t5} & \textbf{Ave.} & \textbf{t1} & \textbf{t2} & \textbf{t3} & \textbf{t4} & \textbf{t5} & \textbf{Ave.} & \textbf{t1} & \textbf{t2} & \textbf{t3} & \textbf{t4} & \textbf{t5} & \textbf{Ave.} \\ 
        \midrule
        Aya-101 & 47.7 & \underline{49.9} & 45.7 & \textbf{51.5} & 48.3 & ${48.6}_{\pm 2.2}$ & 29.6 & \textbf{29.7} & \textbf{29.7} & \textbf{29.7} & 29.6 & ${29.7}_{\pm 0.1}$  & \textbf{4.4} & 4.1 & 4.2 & \textbf{4.4} & 4.3 & ${4.3}_{\pm 0.1}$ \\ 
        Gemma 2 27B & 38.3 & \textbf{40.3} & 33.4 & 34.1 & 35.4 & ${36.3}_{\pm 2.9}$ & \textbf{39.9} & \underline{39.2} & 38.7 & 39.2 & 39.1 & ${39.2}_{\pm 0.4}$ & 22.3 & 26.3 & 25.0 & \textbf{27.0} & 26.2 & ${25.4}_{\pm 1.9}$ \\ 
        GPT-4o & 55.3 & 49.6 & \textbf{64.3} & \underline{62.9} & 59.7 & ${58.4}_{\pm 6.0}$ & \textbf{60.0} & 55.8 & 59.0 & 39.6 & \underline{59.6} & ${54.8}_{\pm 8.7}$ & 46.7 & \textbf{49.8} & 48.0 & 49.5 & \underline{49.6} & ${48.7}_{\pm 1.3}$ \\ 
    \bottomrule
    \end{tabular}
    }
    \vspace{-2mm}
    \caption{\textbf{Ablation results:} Effect of using different prompts (t1, t2, t3, t4, and t5) for \iroko datasets. Best prompt results are in \textbf{bold}. Average computed on only African languages. Second best prompt are \underline{underlined}. } 
    \vspace{-2mm}
\label{tab:ablation_prompts}
\end{table*}

\subsection{Task-specific results}
Here, we examine the individual language performance per task, comparing which task prefers \textit{in-language} \vs \textit{translate-test} evaluation. We compare the performance on a subset of LLMs: \aya, \gemmaTwoL \llamal, and \gpto. Other LLMs are in Appendix \ref{app:task_specific}. 

\paragraph{\ammlu evaluation is better when prompting in-language for closed LLMs} 
\autoref{tab:afrimmlu_table} shows the result of different LLMs on  \ammlu using \textit{in-language} vs. \textit{translate-test}. For \gpto, we found \textit{in-language} prompting to be generally better. 
However, for open models like \llamal and \gemmaTwoL, we find a large improvement in the performance of the \textit{translate-test} for 15 out of 16 African languages. The only language that did not improve is \texttt{wol}, probably due to poor MT performance on NLLB-200, as reported in \citet{nllbteam2022language}. This shows the benefit of translate-test for prompting English-centric LLMs when evaluating low-resource languages. 

\paragraph{\ammlu performance by subjects}
\autoref{tab:mmlu_subjects} shows the result of \gpto by subjects. Interestingly, \textit{elementary math} achieved the best overall accuracy, where 13 out of 16 African languages achieved at least 70\% despite struggling with AfriMGSM. The performance difference to \amath may be due to the multi-choice options of MMLU, which may be slightly easier than free-form answer.  \textit{International law} also achieved impressive performance with 10 out of 16 languages achieving 70\%. All languages struggle the most with \textit{Global facts} including English and French. Similarly, many African languages find it difficult to answer questions in \textit{geography} and \textit{microeconomics} subjects. This presents an opportunity for improving LLMs for the education domain.

\paragraph{\amath performance receives significant boost with translate-test for open models} 
\autoref{tab:afrimgsm_detailed} shows that we can achieve a significant boost in performance with the translate-test on all languages we evaluated. We hypothesize that the current LLMs are better at reasoning in English than other languages. \gemmaTwoL and \llamal improved by $+17.6$ and $+21.0$  respectively when questions are asked in English, while closed models like \gpto dropped in performance.


\subsection{Few-shot results, Cross-Lingual Transfer, Sensitivity to Prompt Templates}

\paragraph{Cross-lingual transfer in-language achieves better results} When there is large enough labeled data in English, we could leverage this cross-lingual signal for zero-shot evaluation. We trained on 400k English NLI examples, and we performed zero-shot transfer in in-language and translate-test setting. \autoref{tab:nli_results} shows that cross-lingual transfer using an Africa-centric smaller language model (AfroXLMR-76L) gave better results than prompting LLMs on average. AfroXLMR-76L has been pre-trained on all languages in AfriXNLI, which explains the impressive performance. However, for multilingual encoders that have not seen some of the languages, prompting \gpto seems to be better, as shown in Appendix \ref{app:cross_tranf}. 

\paragraph{Impact of few-shot vs zero-shot} \autoref{fig:few_shot} shows the few-shot results for the \iroko datasets leveraging \gemmaTwoL, \llamal and \gpto when we provide few examples in in-language setting. 
We found out that \gemmaTwoL and \llamal LLM consistently benefited the least from additional few shots examples for classification tasks (\anli and \ammlu) where \gemmaTwoL improved by $+13.2$ and $+4.9$ respectively. Similarly, \llamal improved on both \anli and \ammlu tasks with $+10.7$ and $+7.0$. However, for reasoning tasks, only \gpto improved in performance by $6.3$, other LLMs dropped in performance, probably due to their inability to reason in non-English languages. Surprisingly, \gpto did not benefit from additional examples for the classification tasks.




\begin{table}[t]
\centering
\scalebox{0.6}{
\begin{tabular}{p{120mm}}
\toprule
 \textbf{Prompt t2} \\
\texttt{\{\{premise\}\}} \\
  \\
\texttt{Question: \{hypothesis\} True, False, or Neither?} \\
 \texttt{Answer:}\\
\midrule
\textbf{Prompt t3} \\
  \texttt{Given the following premise and hypothesis in English, identify if the premise entails, contradicts, or is neutral towards the hypothesis. Please respond with exact 'entailment', 'contradiction', or 'neutral'. } \\
  \\
\texttt{Premise: \{premise\}} \\
 \texttt{Hypothesis:  \{hypothesis\}}\\
\midrule
\textbf{Prompt t4} \\
\texttt{You are an expert in Natural Language Inference (NLI) specializing in the \{Language\} language.
Analyze the premise and hypothesis given in \{Language\}, and determine the relationship between them. 
Respond with one of the following options: ‘'entailment', 'contradiction', or 'neutral'. 
} \\
  \\
\texttt{Premise: \{premise\}} \\
 \texttt{Hypothesis:  \{hypothesis\}} \\
 \bottomrule
\end{tabular}
}
\caption{Three different prompts preferred by different models for AfriXNLI}
\label{tab:xnli_prompts_3}
\vspace{-2mm}
\end{table}

\paragraph{Sensitivity to prompt templates} 
To understand whether sensitivity to prompts impacts results, we perform an ablation for all the \iroko tasks where we evaluate the performance of five different prompts (see \autoref{app:eval_tool}). \autoref{tab:ablation_prompts} shows the results of five prompts we tested for (three of the prompts most preferred by different models are shown in \autoref{tab:xnli_prompts_3}). On \anli, we find that simpler prompt~\citep{nie-etal-2020-adversarial} i.e. \texttt{\{\{premise\}\} Question:
\{\{hypothesis\}\} True, False, or Neither?}, have better results for the open models like \aya and \gemmaTwoL, while \gpto prefers \texttt{t3} where a detailed task description is provided. The best prompt for \aya is \texttt{t4} where the \texttt{\{\{language\}\}} name is mentioned, which shows additional language information may be useful in improving performance. 

On \ammlu, we find \gpto perform worse for \texttt{t4}.\footnote{Analyze each question critically and determine the most correct option based on your understanding of the subject matter '' Question: \{question\}.  Choices A: \{choice1\}, B:\{choice2\}, C: \{choice3\}, D: \{choice4\} }. 
However, other models are not very sensitive to the use of different prompts. In general, we do not find \amath to be sensitive to different prompts. In \autoref{app:task_specific}, we provide the results of five prompt templates for all LLMs evaluated.

\section{Conclusion}
In this paper, we introduced \iroko, a new benchmark for evaluating large language models (LLMs) on African languages. \iroko comprises three datasets focused on different tasks: natural language inference (\anli), multi-choice knowledge QA (\ammlu), and mathematical reasoning (\amath). Unlike previous benchmarks, which primarily involve simple text classification tasks, these datasets assessed the LLMs' abilities in complex and knowledge-intensive areas. Our evaluation revealed a significant performance gap between high-resource languages (\eg English and French) and African languages. Additionally, we observed a substantial disparity in performance between open models and proprietary models, with the latter generally outperforming the former, particularly in mathematical reasoning tasks. We hope that \iroko will serve as a valuable benchmark for evaluating future LLMs developed or adapted for African languages.

\paragraph{Limitations} \label{limitations} Our benchmark has a few limitations: (1) \textbf{The benchmark is human-translated} which may include some translationese effects, it would have been better if they are all generated in the native African languages. However, this parallel translation allows us to evaluate and compare the same sentences in all these languages. (2) \textbf{We only cover three language families in Africa}, Nilo-Saharan, 	
Austronesian, and Khoisan language groups are missing, one of the reason we excluded them is either lack of contact with professional translators or limited translation budget, we hope to extend to more languages in the future. 

\section*{Acknowledgment}

This work was supported by Lacuna Fund, an initiative co-founded by The Rockefeller Foundation, Google.org, and Canada’s International Development Research Centre. Additional support was provided through compute credits from Oracle and the Cohere For AI Research Grant. We extend our gratitude to Charles Riley for facilitating our connection with the Vai translator for the MGSM data translation. 

We are also deeply thankful to OpenAI for granting API credits through their Researcher Access API program to Masakhane, enabling the evaluation of GPT-3.5 and GPT-4 LLMs. Similarly, we appreciate Google for providing GCP credits via the Gemma 2 Academic Program, which supported the Gemini-1.5-Pro inference. Lastly, we would like to thank Hailey Schoelkopf and Lintang Sutawika for their invaluable assistance with the EleutherAI \textit{lm-eval} tool.


\bibliography{custom}

\appendix

\section{Appendix}
\label{sec:appendix}

\begin{table*}[b]

\begin{center}
\scalebox{0.65}{
\begin{tabular}{lllrrrr}
\toprule
\textbf{Language}
& \textbf{Family/branch}
& \textbf{Region}
& \textbf{\# speakers}  & \textbf{\# chars in MADLAD (MB)} & \textbf{In Aya} & \textbf{In BLOOMZ}\\
\midrule
English (eng) & Indo-European / Germanic & Across Africa  & 1457M   & 9,000,000MB & \ding{51} & \ding{51}\\
French (fra) & Indo-European /Romance & Across Africa  & 310M & 1,000,000MB & \ding{51} & \ding{51}\\
Kiswahili (swa) & Niger-Congo / Bantu & East \& Central Africa  & 71M-106M  & 2,400MB& \ding{51} & \ding{51} \\
Kinyarwanda (kin) &  Niger-Congo / Bantu & East Africa    & 10M  & 749MB & \ding{51} & \ding{51} \\
Hausa (hau) & Afro-Asiatic / Chadic & West Africa    & 77M  & 630MB & \ding{51} & \ding{56} \\
Amharic (amh)  & Afro-Asiatic / Ethio-Semitic & East Africa    & 57M &  509MB & \ding{51} & \ding{56}\\
isiXhosa (xho) & Niger-Congo / Bantu & Southern Africa  & 19M &  287MB & \ding{51} & \ding{51}\\
chiShona (sna) & Niger-Congo / Bantu & Southern Africa   & 11M  &  266MB & \ding{51} & \ding{51}\\
isiZulu (xho) & Niger-Congo / Bantu & Southern Africa  & 27M  &  257MB & \ding{51} & \ding{51} \\
Igbo (ibo) & Niger-Congo / Volta-Niger & West Africa    & 31M  & 251MB & \ding{51} & \ding{51} \\
\yoruba (yor) & Niger-Congo / Volta-Niger & West Africa  & 46M & 239MB & \ding{51} & \ding{51}\\
Sesotho (sot) & Niger-Congo / Bantu & Southern Africa  & 13M & 227MB & \ding{51} & \ding{51} \\
Oromo (orm) & Afro-Asiatic / Cushitic & East Africa  & 37M & 88MB & \ding{56} & \ding{56} \\
Luganda (lug) &  Niger-Congo / Bantu & Central Africa    & 11M & 48MB & \ding{56} & \ding{51} \\
Ewe (ewe) & Niger-Congo / Kwa & West Africa    & 7M  &  33MB & \ding{56} & \ding{51} \\
Twi (twi) & Niger-Congo / Kwa & West Africa  & 9M  & 25MB & \ding{51} & \ding{51} \\
Lingala (lin) &  Niger-Congo / Bantu & Central Africa    & 40M   & 22MB & \ding{56} & \ding{51} \\
Wolof (wol) & Niger-Congo / Senegambia & West Africa  & 5M  & 5MB & \ding{56} & \ding{51}\\
Vai (vai) & Mande & West Africa  & 140,000  & - & \ding{56} & \ding{56}\\

\bottomrule
\end{tabular}
}
\caption{\textbf{Languages covered in \iroko}: including language family, region,  number of  L1 \& L2 speakers, size of monolingual data on the web (in MADLAD corpus)}
\label{tab:languages}
\vspace{-4mm}

\end{center}
\end{table*}

\subsection{Language covered}
\label{app:languages}
\autoref{tab:languages} provide the languages covered in the \iroko, their language family, regions located in Africa, number of speakers, and size of monolingual data available on the web based on the MADLAD cleaned corpus~\citep{kudugunta2023madlad400}---we only report number of characters in mega bytes. Additionally,  we added an indication whether this language is covered in the pre-training of \aya and \bloom LLMs. 

\subsection{Prompts Template and Evaluation Tool}
\label{app:eval_tool}
We use the EleutherAI LM Evaluation Harness (\texttt{lm-eval}) tool~\citep{biderman2024lessons}---a popular evaluation tool that is helping to standardize LLM evaluation.  
The tool allows for three types of evaluation: \textit{log-likelihood}, \textit{perplexity}, and \textit{generation}. The log-likelihood is more suitable for multiple-choice tasks since it helps to restrict the model's option to fewer choices---more appropriate for weaker models. However, the log-likelihood approach cannot evaluate the generative capabilities of LLMs to generate coherent and relevant answers. Moreover, closed models are only accessible via API and do not provide access to the log probabilities, making it impossible to use the log-likelihood approach. To extract the correct answers for the task, we employed a \textbf{verbalizer}~\cite{gao-etal-2021-making,schick2021s}. 
For \amath, we used the default verbalizer provided by the tool. However, for \anli and \ammlu, we manually created a verbalizer for the closed models and used the \textit{log-likelihood} request type for the open models.

The prompt templates used for evaluation of different tasks are in \autoref{tab:xnli_prompts}, \autoref{tab:mmlu_prompts} and \autoref{tab:mgsm_prompts}. 



\begin{table*}[t]

\centering
\scalebox{0.80}{
\begin{tabular}{p{200mm}}
\toprule
\textbf{Prompt1} \\
\texttt{Question: \{question\} } \\
 \texttt{Answer: }\\
 \midrule
 \textbf{Prompt2} \\
 \texttt{Give direct numerical answers for the question provided.} \\
\texttt{Question: \{question\} } \\
 \texttt{Answer: }\\
 \midrule
 \textbf{Prompt3} \\
 \texttt{Solve the following math question.} \\
\texttt{Question: \{question\} } \\
 \texttt{Answer: }\\
 \midrule
 \textbf{Prompt4} \\
 \texttt{Answer the given question with the appropriate numerical value, ensuring that the response is clear and without any supplementary information.} \\
\texttt{Question: \{question\} } \\
 \texttt{Answer: }\\
  \midrule
 \textbf{Prompt5} \\
 \texttt{For mathematical questions provided in {language} language. Supply the accurate numeric answer to the provided question
} \\
\texttt{Question: \{question\} } \\
 \texttt{Answer: }\\
 \bottomrule
\end{tabular}}
\caption{Five different prompt used for prompt sensitivity experiments in AfriMGSM}
\label{tab:mgsm_prompts}
\vspace{-2mm}
\end{table*}

\begin{table*}[t]
\centering
\scalebox{0.7}{
\begin{tabular}{p{200mm}}
\toprule
\textbf{Prompt 1} \\
\texttt{Please identify whether the premise entails or contradicts the hypothesis in the following premise and hypothesis. The answer should be exact entailment, contradiction, or neutral.}\\
  \\
\texttt{Premise: \{premise\} } \\
 \texttt{Hypothesis: \{hypothesis\}}\\
 \texttt{Is it entailment, contradiction, or neutral?} \\
 \midrule
 \textbf{Prompt 2} \\
\texttt{\{\{premise\}\}} \\
  \\
\texttt{Question: \{hypothesis\} True, False, or Neither?} \\
 \texttt{Answer:}\\
\midrule
\textbf{Prompt 3} \\
  \texttt{Given the following premise and hypothesis in English, identify if the premise entails, contradicts, or is neutral towards the hypothesis. Please respond with exact 'entailment', 'contradiction', or 'neutral'. } \\
  \\
\texttt{Premise: \{premise\}} \\
 \texttt{Hypothesis:  \{hypothesis\}}\\
\midrule
\textbf{Prompt 4} \\
\texttt{You are an expert in Natural Language Inference (NLI) specializing in the \{Language\} language.
Analyze the premise and hypothesis given in \{Language\}, and determine the relationship between them. 
Respond with one of the following options: ‘'entailment', 'contradiction', or 'neutral'. 
} \\
  \\
\texttt{Premise: \{premise\}} \\
 \texttt{Hypothesis:  \{hypothesis\}} \\
\midrule
\textbf{Prompt 5} \\
\texttt{Based on the given statement, is the following claim 'true', 'false', or 'inconclusive'. } \\
  \\
\texttt{Premise: \{premise\}} \\
 \texttt{Hypothesis:  \{hypothesis\}} \\
 \bottomrule
\end{tabular}
}
\caption{Five different prompt used for prompt sensitivity experiments in AfriXNLI}
\label{tab:xnli_prompts}
\vspace{-2mm}
\end{table*}

\begin{table*}[t]
\centering
\scalebox{0.7}{
\begin{tabular}{p{200mm}}
\toprule
\textbf{Prompt 1} \\
\texttt{You are a highly knowledgeable and intelligent artificial intelligence  model answers multiple-choice questions about \{subject\}}\\
  \\
\texttt{Question: \{question\} } \\
 \texttt{ A: \{choice1\}}\\
 \texttt{ B: \{choice2\}}\\
 \texttt{ C: \{choice3\}}\\
 \texttt{ D: \{choice4\}}\\
 \texttt{Answer:} \\
 \midrule
 \textbf{Prompt 2} \\
\texttt{As an expert in \{subject\}, choose the most accurate answer to the question below. Your goal is to select the correct option 'A', 'B', 'C', or 'D' by understanding the nuances of the topic.}\\
  \\
\texttt{Question: \{question\} } \\
 \texttt{ A: \{choice1\}}\\
 \texttt{ B: \{choice2\}}\\
 \texttt{ C: \{choice3\}}\\
 \texttt{ D: \{choice4\}}\\
 \texttt{Answer:} \\
 \midrule
 \textbf{Prompt 3} \\
\texttt{You are a subject matter expert in \{subject\}. Utilizing your expertise in \{subject\}, answer the following multiple-choice question
  by picking 'A', 'B', 'C', or 'D'.}\\
  \\
\texttt{Question: \{question\} } \\
 \texttt{ A: \{choice1\}}\\
 \texttt{ B: \{choice2\}}\\
 \texttt{ C: \{choice3\}}\\
 \texttt{ D: \{choice4\}}\\
 \texttt{Answer:} \\
 \midrule
 \textbf{Prompt 4} \\
\texttt{Analyze each question critically and determine the most correct option based on your understanding of the subject matter} \\
  \\
 \texttt{ A: \{choice1\}}\\
 \texttt{ B: \{choice2\}}\\
 \texttt{ C: \{choice3\}}\\
 \texttt{ D: \{choice4\}}\\
 \texttt{Answer:} \\
 \midrule
 \textbf{Prompt 5} \\
\texttt{Given your proficiency in \{subject\}, please answer the subsequent multiple-choice question with 'A', 'B', 'C', or 'D'.}\\
  \\
\texttt{Question: \{question\} } \\
 \texttt{ A: \{choice1\}}\\
 \texttt{ B: \{choice2\}}\\
 \texttt{ C: \{choice3\}}\\
 \texttt{ D: \{choice4\}}\\
 \texttt{Answer:} \\
 \bottomrule
\end{tabular}
}
\caption{Five different prompt used for prompt sensitivity experiments in AfriMMLU} 
\label{tab:mmlu_prompts}
\vspace{-2mm}
\end{table*}

\begin{figure*}[t]
\centering
    \begin{subfigure}{0.32\textwidth}
        \centering
\includegraphics[width=0.95\textwidth]{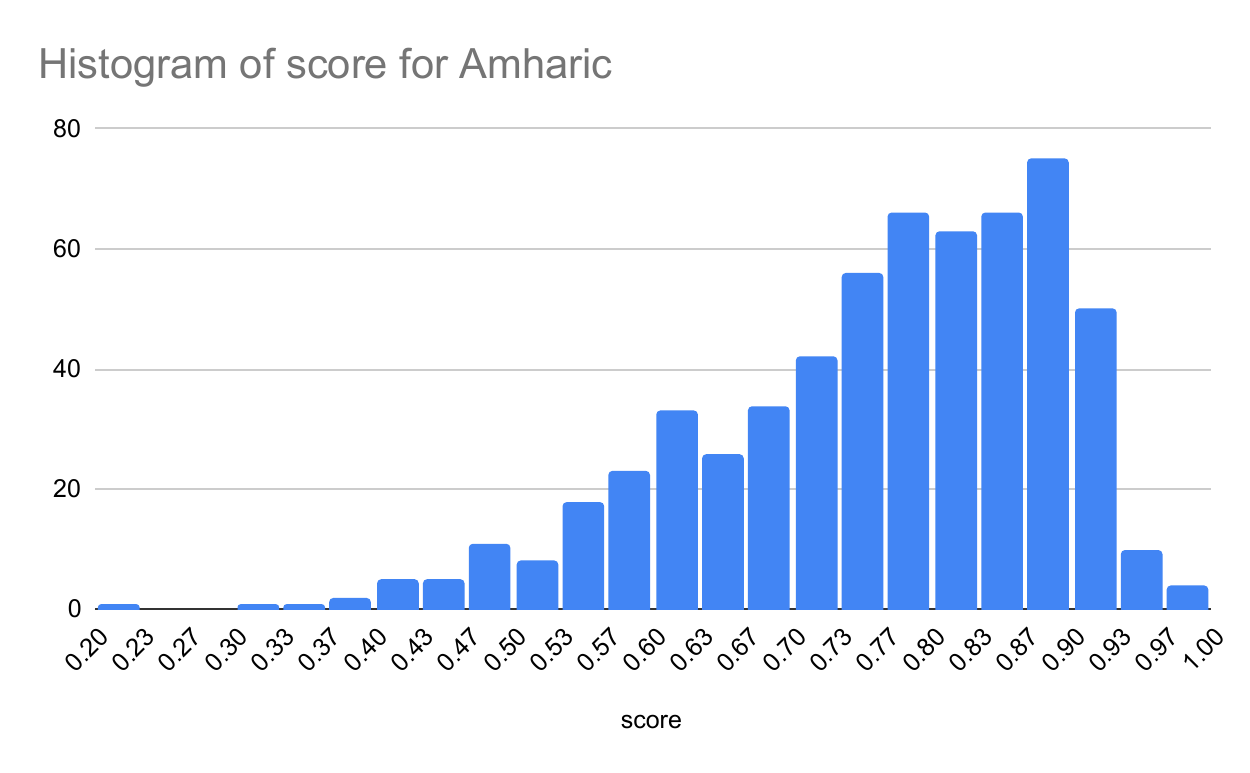}
        \caption{Amharic} \label{fig:amh_comet}
    \end{subfigure}
    \begin{subfigure}{0.32\textwidth}
        \centering
    \includegraphics[width=0.95\textwidth]{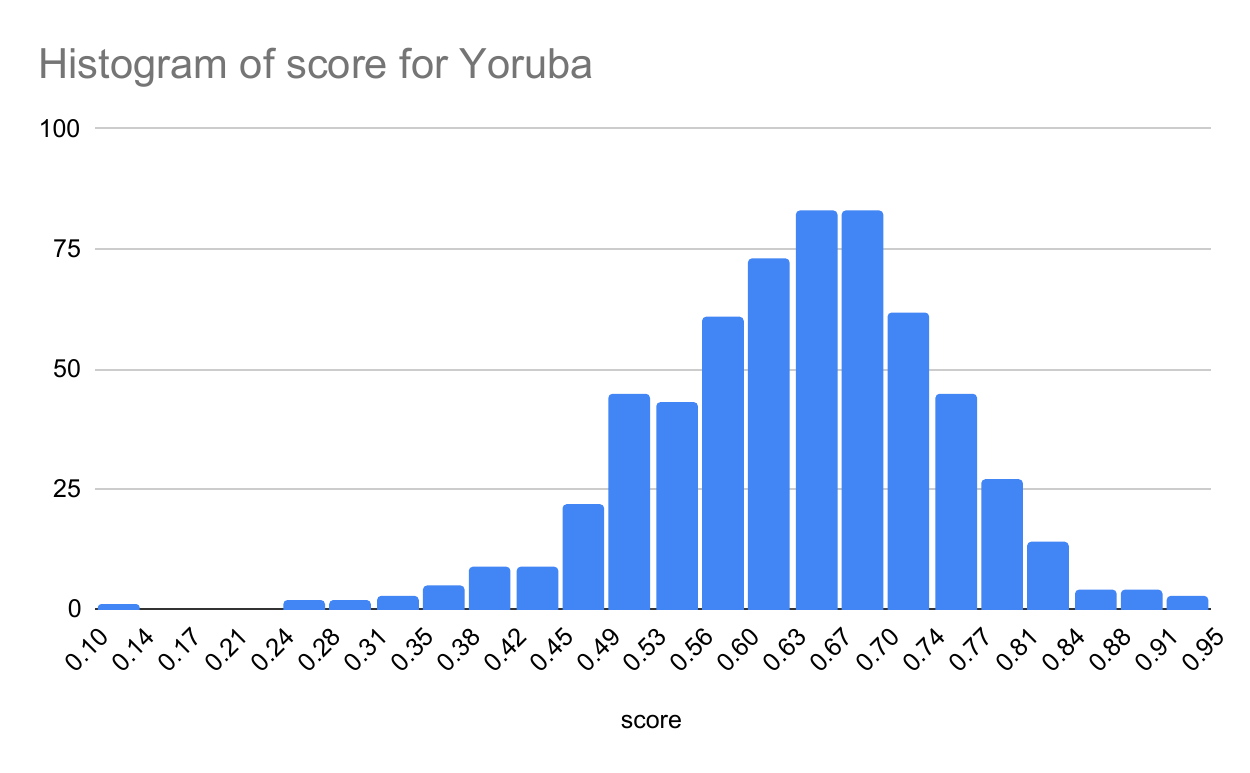}
        \caption{\yoruba} \label{fig:yor_comet}
    \end{subfigure}
    \begin{subfigure}{0.32\textwidth}
        \centering
        \includegraphics[width=0.95\textwidth]{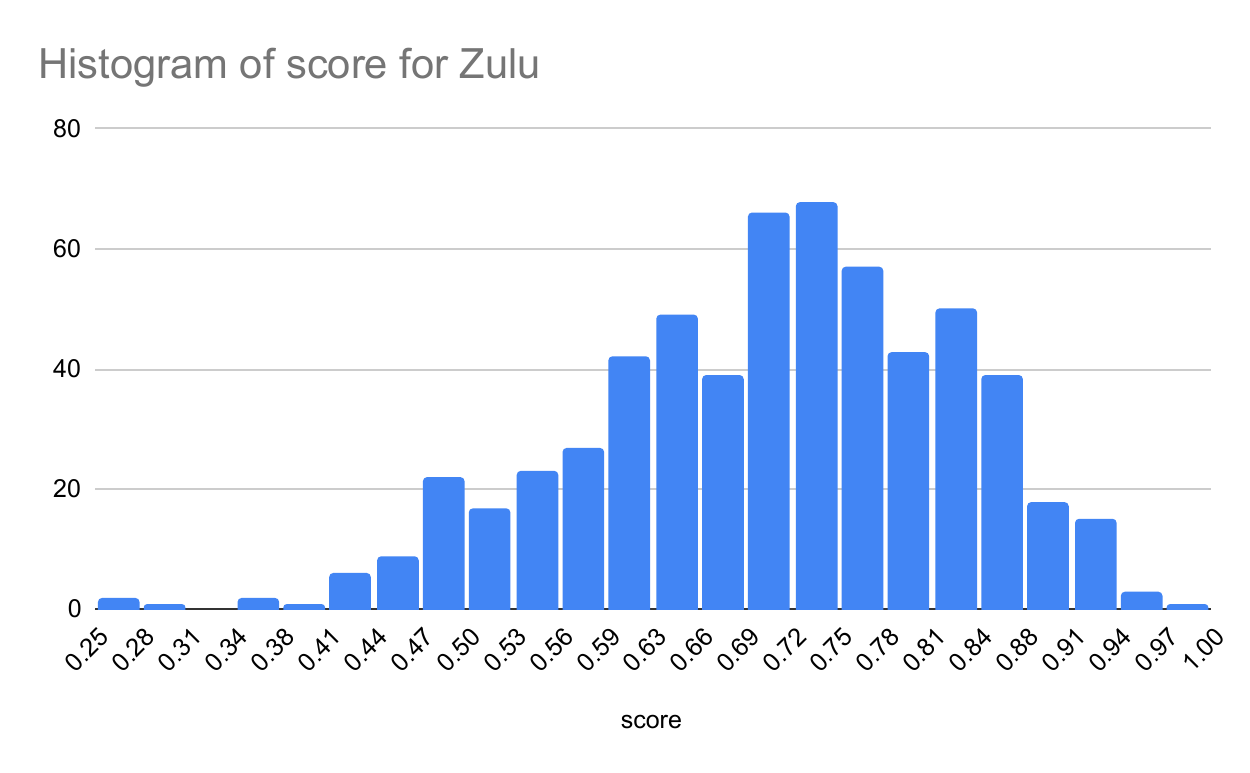}
        \caption{isiZulu} \label{fig:zul_comet}
    \end{subfigure}
    \begin{subfigure}{0.32\textwidth}
        \centering
\includegraphics[width=0.95\textwidth]{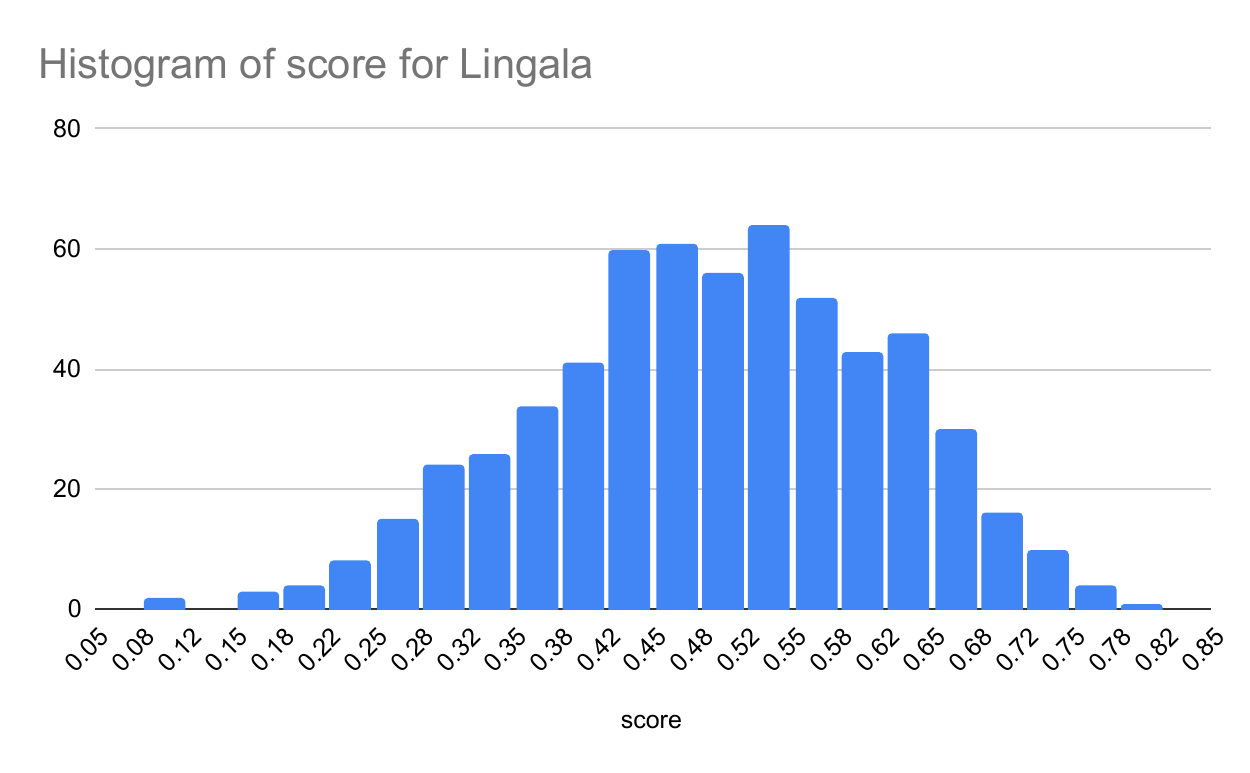}
        \caption{Lingala} \label{fig:lin_comet}
    \end{subfigure}
    \begin{subfigure}{0.32\textwidth}
        \centering
\includegraphics[width=0.95\textwidth]{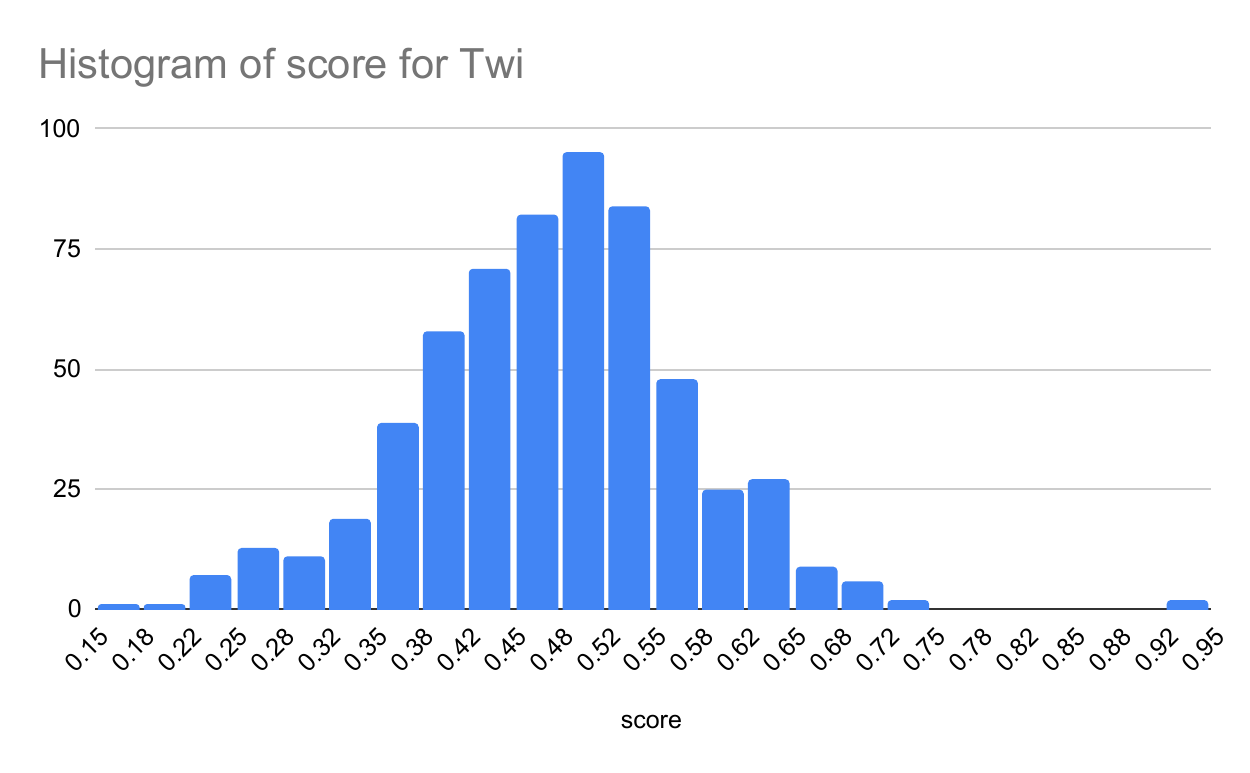}
        \caption{Twi} \label{fig:twi_comet}
    \end{subfigure}
    \begin{subfigure}{0.32\textwidth}
        \centering
\includegraphics[width=0.95\textwidth]{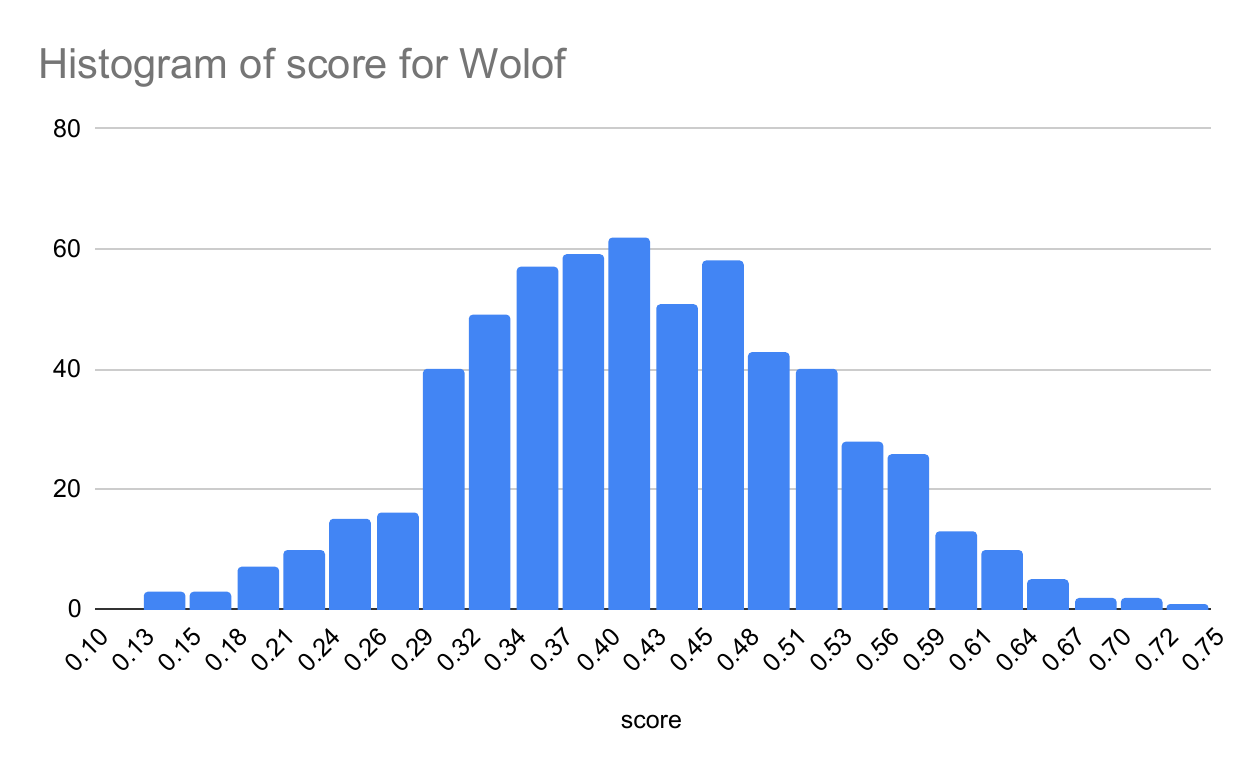}
        \caption{Wolof} \label{fig:wol_comet}
    \end{subfigure}
    
    \caption{Evaluation of \textbf{\anli} translations using AfriCOMET metric scores.
    \label{fig:afrixnli_comet}}
\end{figure*}

\subsection{AfriCOMET metric scores for XNLI translation}
\label{app:africomet_scores}

We employ AfriCOMET evaluation metrics, as developed by \citet{wang-etal-2024-afrimte}, to automatically assess the quality of translations for our newly created benchmarks. Figure \ref{fig:afrixnli_comet} depicts the histogram of scores obtained from AfriCOMET for \anli, illustrating promising results and offering compelling evidence for the effectiveness of our translations (Amharic, \yoruba, isiZulu). However, the performance of this metric depends on if the language we are evaluating is covered in the pre-training of the base model of the metric i.e. AfroXLMR-large. In the case of Lingala, Twi and Wolof, the performance of the metric does not correlate with the human translation since they are not covered in AfroXLMR. Similar findings were reported in the original AfriCOMET QE paper that Twi had worse correlation with human judgement (\ie 0.279 for Pearson, and 0.060 for Spearman) \citep{wang-etal-2024-afrimte}.

\subsection{Task-specific results for all models}
\label{app:task_specific}
We provide the entire results of all LLMs and all prompts on \ammlu, \anli and \amath tasks are shown in \autoref{tab:mmlu_all_models},  \autoref{tab:xnli_all_models} and \autoref{tab:mgsm_all_models}. We performed evaluation on 5 prompts for all models except Claude Opus which we limit to one prompt due to API inference cost.

\subsection{Comparison between in-language and translate-test results}
\label{app:comparison_tt}
We provide the results comparing the in-language and translate-test results on all LLMs in \autoref{tab:xnli_inlang_tt}, \autoref{tab:mmlu_inlang_tt}, and \autoref{tab:mgsm_inlang_tt}.

\subsection{Cross-lingual transfer results for XNLI}
\label{app:cross_tranf}
In \autoref{tab:xlingual_bert}, we compare different multilingual masked language model (MLM) performance on African languages. XLM-R-large has 559M parameters and is trained on 100 languages, but only a few African languages are covered (\texttt{amh}, \texttt{hau}, \texttt{orm}, \texttt{swa}, and \texttt{xho}). Serengeti, on the other hand, has been pre-trained on all languages in \iroko, but it only has 240M parameters. AfroXLMR was adapted from XLM-R through continual pre-training on 17 African languages including 11 in \iroko (\texttt{amh}, \texttt{hau}, \texttt{ibo}, \texttt{kin}, \texttt{orm}, \texttt{sna}, \texttt{sot}, \texttt{swa}, \texttt{xho}, \texttt{yor}, and \texttt{zul}). AfroXLMR-76L follows the same technique by performing continual pre-training on XLM-R-large on 76 languages (72 African), all languages covered in \iroko are part of its pre-training. 

We found Africa-centric MLM to perform better on average than massively multilingual models like XLM-R-large. Serengeti and AfroXLMR-base improved over larger-sized XLM-R-large by $+3.5$ and $+5.3$ points, respectively. Similarly, fine-tuning AfroXLMR-large, a larger version of AfroXLMR-base, results in an improved boost in performance with $11.8$ points. The best overall results were achieved by AfroXLMR-76L with a $16.1$ boost in performance over XLM-R-large. This is probably because all the languages are used in pre-training. We make use of AfroXLMR-76L has the baseline for all LLMs. Interestingly, we find \gpto to be competitive or better than other MLMs except the AfroXLMR-76L on average.

\onecolumn

\begin{scriptsize}\setlength\tabcolsep{3pt}
    \captionsetup{width=15cm}
    \centering

    }
    \caption{\textbf{\anli results for in-language and translate-test}. We make use of the best prompt}
\label{tab:xnli_inlang_tt}
\end{table*}

\begin{table*}[!ht]
    \centering
     \resizebox{\textwidth}{!}{%
     \begin{tabular}{l|rrrrrrrrrrrrrrrrrr|r}
    \toprule
\textbf{Model} & \textbf{eng}  & \textbf{fra} & \textbf{amh}  & \textbf{ewe}  & \textbf{hau}  & \textbf{ibo}  & \textbf{kin}  & \textbf{lin}  & \textbf{lug}  & \textbf{orm}  & \textbf{sna}  & \textbf{sot}  & \textbf{swa}  & \textbf{twi}  & \textbf{wol}  & \textbf{xho}  & \textbf{yor}  & \textbf{zul}  & \textbf{ave}  \\
\midrule
        \textbf{in-language} & ~ & ~ & ~ & ~ & ~ & ~ & ~ & ~ & ~ & ~ & ~ & ~ & ~ & ~ & ~ & ~ & ~ & ~ & ~ \\ 
        mT0-XXL-MT (t1) & 36.4 & 33.0 & 27.4 & 24.8 & 29.2 & 28.4 & 27.2 & 28.6 & 26.0 & 27.4 & 26.6 & 30.4 & 32.2 & 30.2 & 24.8 & 26.8 & 28.8 & 28.2 & 27.9 \\ 
        Aya-101 (t2) & 40.0 & 36.6 & 31.6 & 25.4 & 33.4 & 36.8 & 30.8 & 27.8 & 28.0 & 26.2 & 28.2 & 31.8 & 32.2 & 26.8 & 25.2 & 32.0 & 28.4 & 29.8 & 29.7 \\ 
        BLOOMZ 7B (t2) & 33.8 & 32.2 & 19.0 & 23.6 & 24.0 & 24.2 & 27.0 & 23.0 & 21.2 & 23.6 & 24.8 & 24.2 & 27.2 & 25.2 & 26.0 & 24.4 & 24.6 & 22.8 & 24.1 \\ 
        LLaMa 3 8B (t4) & 55.4 & 46.0 & 29.4 & 26.4 & 27.8 & 30.6 & 30.4 & 28.4 & 25.0 & 31.6 & 26.6 & 28.6 & 32.0 & 26.6 & 26.4 & 23.4 & 29 & 27.8 & 28.1 \\ 
        LLaMa 3.1 8B (t1) & 62.8 & 56.0 & 33.2 & 30.6 & 33.6 & 31.4 & 29.2 & 33.0 & 31.8 & 32.4 & 26.8 & 31.8 & 39.2 & 27.0 & 26.8 & 27.2 & 32.4 & 30.8 & 31.1 \\ 
        LLaMaX 3 8B (t5) & 47.2 & 38.0 & 30.2 & 28.0 & 33.4 & 32.0 & 27.8 & 28.4 & 25.2 & 30.0 & 31.6 & 29.4 & 35.6 & 25.8 & 27.0 & 28.0 & 29.8 & 26.6 & 29.3 \\
\bottomrule
    \end{tabular}
}
\caption{\textbf{\ammlu results for in-language and translate-test}. We make use of the best prompt}
\label{tab:mmlu_inlang_tt}
\end{table*}

\begin{table*}[!ht]
    \centering
    \resizebox{\textwidth}{!}{%
     \begin{tabular}{l|rrrrrrrrrrrrrrrrrrr|r}
    \toprule
\textbf{Model} & \textbf{eng} & \textbf{fra} &\textbf{vai} & \textbf{amh}  & \textbf{ewe}  & \textbf{hau}  & \textbf{ibo}  & \textbf{kin}  & \textbf{lin}  & \textbf{lug}  & \textbf{orm}  & \textbf{sna}  & \textbf{sot}  & \textbf{swa}  & \textbf{twi}  & \textbf{wol}  & \textbf{xho}  & \textbf{yor}  & \textbf{zul}  & \textbf{ave}  \\
\midrule
        \textbf{in-language} & ~ & ~ & ~ & ~ & ~ & ~ & ~ & ~ & ~ & ~ & ~ & ~ & ~ & ~ & ~ & ~ & ~ & ~ & ~ & ~ \\ 
        mT0-XXL-MT (t1) & 3.6 & 4.8 & 0.8 & 4.4 & 1.6 & 4.8 & 1.2 & 3.6 & 2.0 & 4.0 & 0.8 & 3.2 & 4.0 & 3.6 & 1.2 & 2.0 & 4.8 & 2.0 & 3.2 & 2.9 \\ 
        Aya-101 (t1) & 10.8 & 9.6 & 0.8 & 6.8 & 3.2 & 7.2 & 3.2 & 3.6 & 4.4 & 2.4 & 4.8 & 6.8 & 10.8 & 6.0 & 1.6 & 1.2 & 4.4 & 3.2 & 3.6 & 4.6 \\ 
        BLOOMZ 7B (t4) & 2.8 & 2.0 & 1.2 & 1.6 & 2.4 & 2.8 & 1.6 & 1.6 & 1.6 & 2.4 & 1.6 & 0.8 & 1.6 & 0.8 & 0.8 & 2.4 & 1.6 & 2.0 & 2.0 & 1.7 \\ 
        LLaMa 3 8B (t3) & 59.2 & 44.0 & 1.2 & 2.0 & 2.4 & 6.8 & 4.0 & 3.2 & 2.8 & 3.6 & 2.4 & 2.4 & 2.8 & 16.8 & 1.6 & 3.2 & 0.4 & 4 & 3.2 & 3.9 \\ 
        LLaMa 3.1 8B (t4) & 56.8 & 54.0 & 1.2 & 4.4 & 2.4 & 10.0 & 4.4 & 7.2 & 3.2 & 7.6 & 3.6 & 4.8 & 6.4 & 41.2 & 4.0 & 2.8 & 2.4 & 5.6 & 5.2 & 7.2 \\ 
        LLaMaX 3 8B (t4) & 15.2 & 12.0 & 4.0 & 3.6 & 1.6 & 9.6 & 5.2 & 4.4 & 2.8 & 5.2 & 1.6 & 6.0 & 3.2 & 10.8 & 1.6 & 3.2 & 5.6 & 4.8 & 7.2 & 4.8 \\
\bottomrule
    \end{tabular}
}
\caption{\textbf{\amath results for in-language and translate-test}. We make use of the best prompt}
\label{tab:mgsm_inlang_tt}
\end{table*}

\begin{table*}[t]
\centering
\caption{\textbf{Cross-lingual transfer results on \anli}: We fine-tuned various multilingual encoders on English training data, and evaluated on other languages. Best result per language in \textbf{bold}}
\label{tab:crossling_nli_results}
    \resizebox{\textwidth}{!}{%
\begin{tabular}{l|rrrrrrrrrrrrrrrrrr|r}
\toprule
\textbf{Model} & {\textbf{eng}} & {\textbf{fra}} & {\textbf{amh}} & {\textbf{ewe}} & {\textbf{hau}} & {\textbf{ibo}} & {\textbf{kin}} & {\textbf{lin}} & {\textbf{lug}} & {\textbf{orm}} & {\textbf{sna}} & {\textbf{sot}} & {\textbf{swa}} & {\textbf{twi}} & {\textbf{wol}} & {\textbf{xho}} & {\textbf{yor}} & {\textbf{zul}} & {\textbf{ave}} \\
\midrule
XLM-R-large & \textbf{90.5} & \textbf{84.3} & 75.8 & 37.2 & 68.2 & 39.3 & 40.5 & 32.8 & 37.7 & 59.2 & 41.2 & 39.7 & 74.2 & 39.0 & 43.8 & 63.5 & 38.5 & 62.2 & 49.6\\
Serengeti & 77.3 & 60.7 & 54.8 & 51.3 & 61.0 & 56.0 & 55.2 & 36.2 & 55.3 & 46.3 & 58.3 & 55.0 & 66.2 & 42.7 & 43.8 & 56.3 & 53.5 & 57.8 & 53.1\\
Afro-XLMR-base & 81.5 & 78.5 & 71.3 & 36.3 & 68.8 & 59.8 & 57.7 & \textbf{37.8} & 44.2 & 56.7 & 59.8 & 61.5 & 67.0 & 40.8 & 41.0 & 61.0 & 52.2 & 62.7 & 54.9\\
Afro-XLMR-large & 86.5 & 82.3 & 77.2 & 39.7 & 75.2 & 69.8 & 64.5 & 35.3 & 57.8 & 69.7 & 68.0 & 69.2 & \textbf{74.3} & 39.5 & 39.8 & 69.0 & 61.0 & 72.2 & 61.4 \\
Afro-XLMR-76L-large & 88.2 & 83.3 & \textbf{78.5} & \textbf{58.3} & \textbf{73.3} & \textbf{70.0} & 65.8 & 33.3 & \textbf{68.0} &\textbf{69.3} & \textbf{70.8} & \textbf{70.8} & 73.3 & \textbf{59.5} & \textbf{51.8} & \textbf{73.0} & \textbf{63.2} & \textbf{72.5} & \textbf{65.7}\\
\midrule
GPT-4o & 86.2 & 78.7 & 66.7 & 48.3 & 69.2 & 68.2 & \textbf{66.8} & 31.2 & 67.2 & 66.2 & 69.8 & 68.3 & 72.5 & 53.2 & 49.5 & 72.5 & 63.5 & 70.0 & 62.7 \\
\bottomrule
\end{tabular}
}
\label{tab:xlingual_bert}
\vspace{1mm}
\end{table*}

\end{document}